\documentclass[11pt]{article}

\PassOptionsToPackage{pagebackref,breaklinks}{hyperref}
\usepackage[preprint]{acl}

\usepackage{times}
\usepackage{latexsym}

\usepackage[T1]{fontenc}

\usepackage[utf8]{inputenc}

\usepackage{microtype}

\usepackage{inconsolata}

\usepackage{graphicx}

\usepackage{dsfont}
\usepackage{bbm}
\usepackage{amsmath} 
\usepackage{amsmath, amssymb, dsfont} 
\usepackage{multirow} 
\usepackage{booktabs} 
\usepackage{array}    
\usepackage{booktabs} 

\usepackage[table]{xcolor}  
\usepackage{graphicx}       
\usepackage{algorithm}
\usepackage{algpseudocode}
\usepackage{multirow}
\usepackage{tcolorbox}
\tcbuselibrary{breakable,skins,theorems}
\definecolor{cvprblue}{rgb}{0.21,0.49,0.74}
\definecolor{mygray}{gray}{.9}
\hypersetup{colorlinks,allcolors=cvprblue}
\usepackage{colortbl}
\usepackage{graphicx} 
\usepackage{subcaption} 
\usepackage{float} 

\title{Scaling with Confidence: Calibrating Confidence of LLMs for Adaptive Test Time Scaling}

\author{
  \textbf{Xuqing Yang\textsuperscript{1}},
  \textbf{Yi Yuan\textsuperscript{2}},
  \textbf{Shanzhe Lei\textsuperscript{3}},
  \textbf{Xuhong Wang\textsuperscript{3}\thanks{Correspondence: \href{mailto:wangxuhong@pjlab.org.cn}{wangxuhong@pjlab.org.cn}}}
\\
\\
  \textsuperscript{1}Shanghai Jiao Tong University,
  \textsuperscript{2}Southeast University,
  \textsuperscript{3}Shanghai AI Laboratory
\\
}

\begin{document}
\maketitle
\begin{abstract}
Training large language models (LLMs) with reinforcement learning (RL) has significantly advanced their performance on reasoning and question-answering tasks. However, prevailing RL reward designs typically prioritize response correctness, neglecting to incentivize models to express their confidence accurately. This leads to a critical problem: performance gains are often accompanied by poor calibration between confidence and accuracy, misleading models to overconfidently hallucinate when uncertain. 
To address this limitation, we propose \textbf{C}orrectness and \textbf{C}onfidence \textbf{C}alibration \textbf{R}einforcement \textbf{L}earning (\textbf{C3RL}), a novel RL algorithm integrating correctness, calibration and dataset-informed reference accuracy rewards together. Comprehensive evaluation across 8 text and multimodal datasets demonstrates that C3RL enhances calibration without sacrificing accuracy, outperforming the current state-of-the-art method in both performance and calibration metrics.
Utilizing the well-calibrated verbalized confidence from C3RL, we further introduce \textbf{C}onfidence-based \textbf{A}daptive Test Time \textbf{S}caling (\textbf{CAS}), an adjustable inference-time strategy that allocates computational resources based on response confidence. Experiments show that CAS surpasses majority voting on both in-domain and out-of-domain datasets while reducing the inference budget by up to 12.33 times. We believe the synergy of C3RL and CAS paves the way for deploying more reliable and resource-efficient LLMs.
The code, data and models will be released.

\end{abstract}

\section{Introduction}
Large Language Models (LLMs) have demonstrated remarkable performance in reasoning, question answering, and code generation \citep{brown2020language, achiam2023gpt}. 
However, their impressive capabilities are accompanied by a critical failure mode: when faced with uncertainty, LLMs frequently produce overconfident but incorrect outputs—a phenomenon widely referred to as \textit{hallucination} \citep{ji2023survey}. Such behavior is not merely an academic curiosity, but 
can lead to serious consequences, including misinformation propagation and decision failures in safety-critical domains.

To mitigate these risks, it is desirable for LLMs not only to generate correct outputs but also to reliably express their epistemic uncertainty. This property, known as \textit{confidence calibration}, refers to the ability of a model to produce probability estimates that faithfully reflect the likelihood of correctness \citep{guo2017calibration}. Well-calibrated confidence serves as a foundation 
to abstain from uncertain answering, defer to human oversight, or dynamically adjust computational budget when confidence is low.

Despite its importance, confidence calibration for LLMs remains underexplored and faces several limitations. 
First, they typically reduce calibration to a binary classification task (certain vs. uncertain), which is too coarse to capture the nuanced uncertainty of LLM predictions. Second, they often improve calibration at the expense of accuracy, since models may learn to produce deliberately incorrect low-confidence answers to minimize calibration loss. Moreover, they rarely explore the potential of calibrated confidence for downstream decision-making, such as adaptive inference and resource allocation \citep{zhao2024fact}. 

\textbf{Our work addresses these gaps by jointly optimizing correctness and calibration within a reinforcement learning framework and by exploiting calibrated confidence for adaptive test-time scaling.} Specifically, we design an approach that explicitly teaches LLMs to signal low confidence when they are uncertain
. Moreover, we leverage the resulting calibrated confidence to guide early stopping in test time scaling, adjusting more computation to low-confidence samples. To the best of our knowledge, our method is the first to integrate confidence calibration and test-time scaling in a unified framework, thereby improving both the reliability and efficiency of inference.

Our contributions are summarized as follows:

\begin{enumerate}
\item \textbf{Correctness and Confidence Calibration Reinforcement Learning(C3RL):} We propose C3RL, a novel RL algorithm that simultaneously improves correctness and calibration, preserving overall performance while reducing overconfidence.
\item \textbf{Confidence-based Adaptive Test Time Scaling(CAS):} We introduce an early-stopping mechanism for test time scaling that leverages calibrated confidence to dynamically allocate inference budget. This approach dedicates more computation to low-confidence ones, striking an improved efficiency–accuracy balance.
\item \textbf{Unified Framework:} Together, C3RL and CAS form a coherent pipeline that enables LLMs to express calibrated confidence and to adaptively control inference cost, paving the way for more trustworthy and efficient LLMs.
\end{enumerate}

\section{Related Work}
Recent studies reveal that, LLMs often exhibit  overconfidence~\citep{xiong2023can, mei2025reasoning} and hallucination~\citep{zhang2025siren, liang2024learning}, which undermines their reliability. Existing efforts to improve confidence estimation and calibration can be broadly grouped into two major lines: (i) prompt and sampling-based calibration and (ii) training and RL-based calibration. We further discuss related work on 
adaptive test time scaling.

\subsection{Prompt and Sampling based Calibration} 
Prompt-based approaches guide LLMs to explicitly verbalize their confidence or self-assess their reliability after producing an answer. 
~\citep{tian2023just, xiong2023can, yang2024verbalized} show that verbalized confidence is often poorly calibrated
though calibration improves with model scale and reasoning capability. More recent reasoning-oriented studies~\citep{mei2025reasoning, yoon2025reasoning} demonstrate that introspective or ``slow thinking'' chain-of-thought (CoT) prompting can yield better alignment between confidence and correctness. However, these prompting-based techniques 
depend heavily on prompt wording, limiting robustness and scalability~\cite{zhao2021calibrate}.

Another line of research estimates confidence implicitly from response consistency. These sampling-based approaches rely on response diversity or majority voting~\citep{kuhn2023semantic, aichberger2025improving, kang2025scalable}, where agreement across multiple generations serves as uncertainty. While these methods correlate well with empirical accuracy, they are computationally expensive.


\begin{figure*}[ht]
    \centering
    \includegraphics[trim=20 110 200 100,clip, width=2\columnwidth]{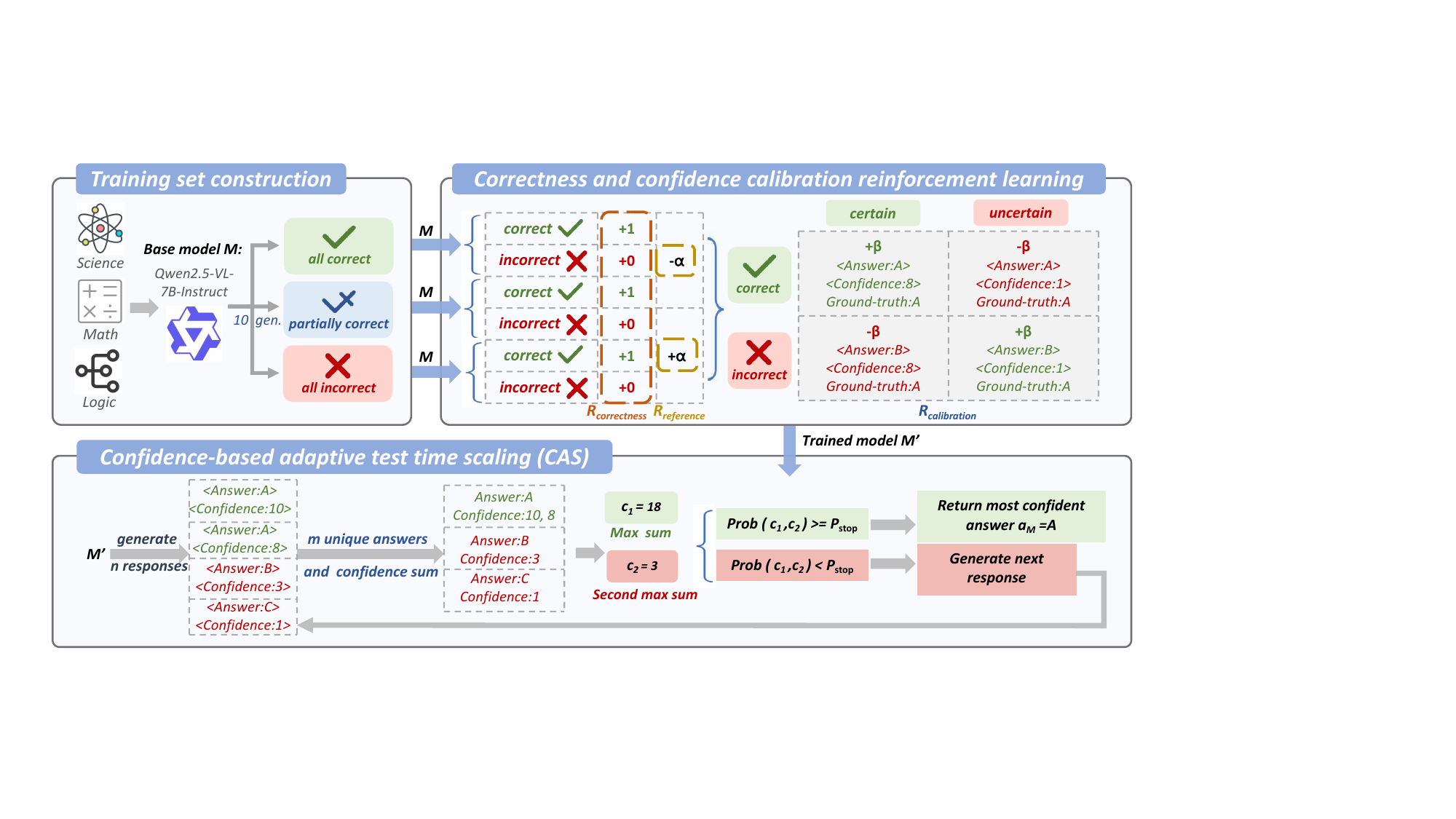}
    \caption{Pipeline for the framework of training set construction, Correctness and Confidence Calibration Reinforcement Learning(C3RL) and Confidence-based Adaptive Test Time Scaling(CAS). The base model first generates 10 responses for each sample and categorizes the samples into ``all correct", ``partially correct" and ``all incorrect". In C3RL, $R_{\mathrm{correctness}}$ encourages correct answers, $R_{\mathrm{reference}}$ rewards/penalizes correctness change based on the assigned categories, and $R_{\mathrm{calibration}}$ incentivizes calibration of confidence. In CAS, the stopping probability (Eq. \ref{equation: probability}) is calculated from the most confident answer and the second most confident answer. If the most confident answer dominates existing generations ($c_1$ is much larger than $c_2$), the inference will stop to save budget.}
    \label{fig:pipeline}
\end{figure*}

\subsection{Training and RL based Calibration}
Training-based methods aim to learn calibrated confidence distributions through supervised or reinforcement objectives. \citet{lin2022teaching} introduces dataset-level supervision by correlating response types with empirical accuracy, while R-Tuning~\citep{zhang2023r} augments supervised fine-tuning data with binary ``I am sure/unsure'' labels. Although effective for coarse uncertainty, such designs fail to capture fine-grained, context-dependent confidence.  

More principled formulations employ reinforcement learning (RL) to jointly optimize correctness and calibration. \textit{SaySelf}~\citep{xu2024sayself} and \citet{stangel2025rewarding} train LMs with proper scoring rules (e.g., Brier or log loss) as rewards, directly aligning verbalized confidence with output correctness. However, these objectives often prioritize calibration alone, potentially encouraging low-confidence incorrect answers to minimize loss. RLCR~\citep{damani2025RLCR} incorporates calibrated reward functions
to balance accuracy and calibration, though applications remain limited to question-answering or Maths tasks. Instead, our model trained with C3RL on reasoning-intensive tasks maintains accuracy and reaches optimal improvement in calibration.


\subsection{Adaptive Test Time Scaling}
Adaptive sampling has emerged as a powerful approach to reduce the computational cost of test time scaling. Adaptive-Consistency~\citep{aggarwal2023AC} introduces a model-agnostic technique that dynamically adjusts the number of samples per question using lightweight stopping criteria based on the frequency of answers. \citet{li2023ESC} takes an alternative approach by dividing the sampling process into sequential windows and stopping when all samples within a window are the same. However, these 
adaptive sampling methods rely on frequency or entropy to approximate confidence, failing to utilize the confidence signal itself directly. 

\section{Method}
In this section, we introduce the construction process of training data (Section \ref{training dataset}), the reward design for C3RL (Section \ref{reward design}) and the adaptive test time scaling strategy based on calibrated confidence (Section \ref{CAS}). Figure~\ref{fig:pipeline} shows the pipeline of the whole framework.

\subsection{Training Dataset Construction}
\label{training dataset}
Training LLMs via Reinforcement Learning requires comprehensive and moderately difficult datasets. 
We create our high-quality training datasets and assign an accuracy tag to each data.

We first collect datasets in mathematics, science and logic reasoning.
The selected  datasets are denoted as $\mathcal{D}$, with the following components: (1)\textbf{ Mathematics and Science:} NuminaMath-TIR~\citep{numina_math} (69k numerical-answer math problems targeting computation and reasoning) and WebInstruct-verified~\citep{webinstruct} (4.5k selected numerical science problems with integer answers); (2)\textbf{ Logical Reasoning:} LogicNLI~\citep{logicnli} (16k problems unbinding the target first-order logic reasoning from commonsense inference) and LogiQA~\citep{logiqa} (8,678 QA problems focusing on deductive reasoning). The detailed filtering process is in~\ref{appendix:training set}. $\mathcal{D}$ is randomly split into $\mathcal{D}_{train}$ (106k samples, 99\%)  and $\mathcal{D}_{test}$ (1,073 samples, 1\%).

For each question $q$ in $\mathcal{D}_{train}$, we use the base model (Qwen2.5VL-7B-Instruct~\cite{qwen2.5-vl} or Llama-3.2-3B-Instruct~\cite{llama3_2_3b}) to generate 10 answers with a decoding temperature 0.7. The answer generation prompt is in~\ref{appendix: answer generation}. 
We calculate the accuracy of this answer set: $Acc_q = \sum_{i=1}^{10} {\mathbb{I}(a_i=a^*)} /{10}$, where $\mathbb{I} \in \{0,1\}$ is an indicator function and $a^*$ is the ground truth. Then we assign a reference accuracy tag $T_q$ from the following three options: ``all correct'' ($Acc_q$ = 1), ``partially correct'' (0<$Acc_q$ <1) and ``all incorrect'' ($Acc_q$ = 0).

To be specific, $D_{train}$ composes of 13.7k ``all correct'' questions, 68.1k ``partially correct'' questions, and 24.3k ``all incorrect'' questions. In the following section, we incorporate this reference accuracy into our reward design.

\subsection{C3RL }
\label{reward design}

Previous RL training methods for calibration simply incentivize calibration and pose potential risks that models learn to give incorrect answers deliberately with low confidence. To prevent such a ``calibrated but wrong" shortcut, we design a novel Correctness and Confidence Calibration Reinforcement Learning (\textbf{C3RL}). 
For each question $q \in \mathcal{D}_{train}$, we prompt the base model to generate an answer $a_q$ with an associated verbalized confidence $c_{a_q} \in \{1,2,...,10\}$, where $c_{a_q}=1$ means the model is totally uncertain and $c_{a_q}=10$ means the model is totally certain. 
We train the model (same as the base model) utilizing the following reward: 
\begin{equation}
\begin{aligned}
    R_{\mathrm{C3RL}}(q,a_q,c_{a_q},T_q)
     = {} & R_{\mathrm{correctness}} 
     + R_{\mathrm{calibration}} \notag \\
     & + R_{\mathrm{reference}} 
\end{aligned}
\end{equation}

The three parts in the reward are as follows:

\textbf{Correctness Reward:} The traditional RLVR~\cite{RLVR} reward $R_\mathrm{{correctness}}(a_q)$ = $\mathbb{I}(a_q=a^*) \in \{0,1\}$. This part ensures the model will update in the ``correct'' direction instead of ``calibrated but wrong'' way.

\textbf{Calibration Reward:} 
We set a confidence threshold $t$=5 to classify the answer $a_q$ into ``certain'' ($c_{a_q}>t$) and ``uncertain'' ($c_{a_q}\leq t$) categories. We encourage ``certain correct'' and ``uncertain incorrect'' answers with $R_{\mathrm{calibration}}=+\beta$ while penalizing ``certain incorrect'' and ``uncertain correct'' responses with $R_{\mathrm{calibration}}=-\beta$. The ablation process to determine $\beta$ is detailed in~\ref{appendix:ablation}.

\textbf{Reference Accuracy Reward:} To utilize the prior accuracy information in data construction process, we customize a reward for question $q$ based on reference accuracy tag $T_q$. When the model answers correctly from formerly ``all incorrect" questions, we incentivize it with $R_{\mathrm{reference}}=+\alpha$. On the contrary, if the model hallucinates with incorrect answers from formerly ``all correct" questions, we punish it with $R_{\mathrm{reference}}=-\alpha$.
This design forces the model to remain correct if it can originally give correct answers and further stops the model from falling into ``calibrated but wrong" risks. The ablation process to select $\alpha$ is elaborated in~\ref{appendix:ablation}.

\subsection{Confidence-based Adaptive Test Time Scaling (CAS)}
\label{CAS}
Prior adaptive self-consistency methods typically rely on frequency or entropy to determine early-stopping criteria. However, such signals are often uncalibrated and fail to accurately reflect the model’s true confidence. To address this limitation, we propose a Confidence-based Adaptive test-time Scaling (CAS) strategy, which integrates calibrated verbalized confidence into the stopping criterion. After generating each batch of responses, CAS applies a  well-calibrated stopping criterion based on the verbalized confidence of model outputs to decide whether to (1) continue sampling or;  (2) halt and return the current highest confidence-weighted answer. 

Inspired by the idea that a high-confidence answer which is dominant over all the other responses suggests a reliable stopping criteria, we design CAS as follows: 

For a question $q$ to be sampled $N$ times, there have been $n<N$ generated responses $\{r_{1}, r_{2},...,r_{n}\}$ with $m$ ($m \leq n$) unique extracted answers $A_q=\{a_1,a_2,...,a_m\}$, and for each unique answer $a_i \in A_q$, assume it corresponds to $j$ responses $R_{a_i}=\{r_{i_1}, r_{i_2},...r_{i_j}\}$. Then we calculate the sum of verbalized confidence for $a_i$: 
$c_{a_i}={1\over{10}}\sum_{k=1}^{j}{\mathrm{conf}(r_{i_k})}$,
where $\mathrm{conf}(r)\in \{1,2,...,10\}$ extracts the verbalized confidence of response $r$. The sums of verbalized confidence for all unique answers in $A_q$ form a confidence set $C_q=\{c_{a_1},c_{a_2},...,c_{a_m}\}$, and we can select the maximum and second maximum of $C_q$: 
    $c_1 = \mathrm{max}(C_q), c_2=\mathrm{max}(C_q \backslash \{c_1\})$.
Intuitively, $c_1$ corresponds to the "most confident" answer, and $c_2$ corresponds to the "second most confident" answer. Following the BETA-distribution stopping criteria described in Adaptive-Consistency~\citep{aggarwal2023AC}, our confidence-based stopping probability can be calculated as:

\begin{equation}
\label{equation: probability}
    \mathrm{Prob}(c_1,c_2) = {1\over A} \int_0^{0.5} x^{c_2}(1-x)^{c_1} \, dx, 
\end{equation}
where coefficient $A=\int_0^{1} x^{c_2}(1-x)^{c_1} \, dx$ normalizes the integral on the interval [0,1] to 1. We set a probability threshold $P_{\mathrm{stop}}$ to decide whether to stop sampling. If $\mathrm{Prob}(c_1,c_2)\geq P_{\mathrm{stop}}$ or $n$ reaches the maximum generation time $N$, the sampling process ends; otherwise, the inference continues with next generation. When $c_1$ is larger compared to $c_2$ (which means the most confident answer dominates the generations), the probability to stop is higher.

\definecolor{codebg}{RGB}{255,255,255}

\begin{algorithm}
\caption{Confidence-based Adaptive Test Time Scaling (CAS)}
\label{alg:confidence_inference}
\begin{minipage}{\linewidth}
\colorbox{codebg}{%
\begin{minipage}{0.98\linewidth}
\renewcommand{\baselinestretch}{0.9}\selectfont
\begin{algorithmic}[1]
\setlength{\itemsep}{0pt}
\Require Question $q$, max sampling time $N$, threshold $P_{\mathrm{stop}}$ 
\State Initialize $n \gets 0$, $A_q \gets \emptyset$, $C_q \gets \emptyset$
\While{$n < N$ \textbf{and} stopping condition not met}
    \State \hangindent=2em \hangafter=1 Generate 1 additional sample, $n \gets n+1$
    \State Update answer set $A_q$, confidence set $C_q$
    \State $c_1, c_2 \gets$ max and second max of $C_q$
    \State Compute $\mathrm{Prob} = \frac{1}{A} \int_0^{0.5} x^{c_2}(1-x)^{c_1} dx$
    \If{$\mathrm{Prob} \geq P_{\mathrm{stop}}$} \textbf{break} \EndIf
\EndWhile
\State \Return $a_M$ (most confident answer)
\end{algorithmic}
\end{minipage}%
}
\end{minipage}
\end{algorithm}
The detailed algorithm is described in Algorithm~\ref{alg:confidence_inference}. We also design a modified equivalent Algorithm~\ref{alg:confidence_inference_parallel} to support parallel inference instead of generating responses one by one. The detail of Algorithm~\ref{alg:confidence_inference_parallel} is elaborated in \ref{parallel CAS}.

\section{Experiments}
The main experiments aim to address the following three research questions: (1) \textbf{how C3RL balances accuracy and calibration} (Sec \ref{section:evaluation of C3RL}); (2) \textbf{how CAS saves inference budget in test times scaling} (Sec \ref{sec: CAS}) and (3) \textbf{how C3RL contributes to CAS performance} (Sec \ref{sec:C3RL-CAS}).
\subsection{Evaluation of C3RL}
\label{section:evaluation of C3RL}

\subsubsection{Benchmarks}
\label{experiments: benchmarks_C3RL}
The benchmarks include text and multimodal datasets across knowledge-intensive tasks: 
(1) \textbf{Text:} AGIEval~\citep{agieval}
and MMLU test set~\citep{mmlu}.
(2) \textbf{Multimodal:} MMMU~\citep{mmmu},  
MathVista~\citep{mathvista} and LogicVista~\citep{logicvista}. The details of the benchmarks are in~\ref{appendix:benchmark}.

\subsubsection{Evaluation metrics}
\begin{table*}[ht]
    \centering
    \renewcommand{\arraystretch}{0.85}
    \setlength{\tabcolsep}{4pt}
    \resizebox{\textwidth}{!}{%
        \begin{tabular}{l|ccc|ccc|ccc|ccc}
            \toprule[1pt]\midrule
                \textbf{Text}
              & \multicolumn{3}{c}{\textbf{In-domain}} 
              & \multicolumn{3}{c}{\textbf{Agieval}} 
              & \multicolumn{3}{c}{\textbf{MMLU}}  
              & \multicolumn{3}{c}{\textbf{Avg.}}  
              \\
            \cmidrule(lr){2-4} \cmidrule(lr){5-7} \cmidrule(lr){8-10} \cmidrule(lr){11-13}
            \textbf{Method} & Acc$\uparrow$ & AUROC$\uparrow$ & ECE$\downarrow$ & Acc$\uparrow$ & AUROC$\uparrow$ & ECE$\downarrow$ & Acc$\uparrow$ & AUROC$\uparrow$ & ECE$\downarrow$ & Acc$\uparrow$ & AUROC$\uparrow$ & ECE$\downarrow$  \\
            \midrule
            \multicolumn{13}{c}{\textbf{Qwen2.5VL-7B-Instruct}} \\
            \midrule
            Base & 45.3\% & 0.531 & 0.448 & 51.0\% & 0.543 & 0.375 & 68.3\% & 0.587 & 0.188 & 54.9\%&	0.554	&0.337 \\
            SC & 53.9\% & 0.762 & 0.088 & 53.7\% & 0.711 & 0.208 & 69.9\% & 0.759 & 0.141 &59.2\%	&0.744	&0.146 \\
            SFT+Ref & 48.0\% & 0.720 &	0.098 & 49.1\%	&0.753&	0.105&	62.1\%	&0.706	&0.098 &53.1\%	&0.726	&0.100
            \\ 
            \midrule
            RLVR & \textbf{66.6\%} & 0.514 & 0.285 &  \textbf{58.1\%} & 0.534 & 0.377 &  71.1\% & 0.602 & 0.239 &\textbf{65.3\%}	&0.550	&0.300 \\
            SaySelf & 55.0\% & \textbf{0.807} & \textbf{0.075} & 50.8\% & \textbf{0.745} & \textbf{0.147} &  64.0\% & \textbf{0.731} & 0.129 &56.6\%	&\textbf{0.761}	&\underline{0.117} \\
            RLCR & \underline{65.1}\% & \underline{0.751} & 0.109 & 57.5\% & 0.664 & 0.205 & 70.7\% & 0.607 & 0.175 &64.4\%	&0.674	&0.163 \\
            C3RL w/o Ref & 64.9\% & 0.700 & 0.119 & \textbf{58.1\%} & 0.682 & 0.197 & \underline{71.2\%} & 0.648 & \underline{0.111} &\underline{64.7\%}	&0.677	&0.142 \\
            \rowcolor[gray]{0.9} \textbf{C3RL (Ours)} & 64.4\% & 0.748 & \underline{0.076} & \underline{57.8\%} & \underline{0.688} & \underline{0.154} & \textbf{71.4\%} & \underline{0.653} & \textbf{0.092} &64.5\%	&\underline{0.696}	&\textbf{0.107} \\
            \bottomrule[1pt]
            \addlinespace[2pt]
            \multicolumn{13}{c}{\textbf{Llama-3.2-3B-Instruct}} \\
            \midrule
            Base & 37.2\%	& 0.563	&0.134 & 34.8\%	&	0.557	&0.423	&59.6\%		&0.586	&0.271 &43.9\%	&0.569	&0.276\\
            SC &  40.8\%	&	0.787&	0.091& 28.6\%	&	0.644	&0.352&	59.9\%	&	0.733	&0.158 &43.1\%	&0.721	&0.200\\
            SFT+Ref & 37.6\%	&0.572	&0.168 &32.1\%	&0.594	&0.240	&52.8\%	&0.628	&0.162 &40.8\%	&0.598	&0.190 \\
            \midrule
            RLVR & 45.9\%	&0.587	&0.463 &\textbf{37.3\%}	&0.522	&0.468	&63.4\%	&0.537	&0.208  &48.9\%	&0.549	&0.380\\
            SaySelf & 13.1\%	&0.676	&\textbf{0.060} &19.4\%		&0.611	&\textbf{0.072}	&37.4\%	&\textbf{0.715}	&0.110 &23.3\%	&0.667	&\textbf{0.081}\\
            RLCR & 43.3\%	&\textbf{0.799}	&\underline{0.072} &34.4\%	&0.658	&\underline{0.111}	&62.3\%	&0.670	&0.153 &46.7\%	&\textbf{0.709}	&\underline{0.112}\\
            C3RL w/o Ref & \underline{47.3\%}	&0.697	&0.136 &\underline{37.2\%}	&0.652	&0.218	&\underline{63.5\%}	&\underline{0.688}	&\textbf{0.087} &\underline{49.3\%}	&0.679	&0.147\\
            \rowcolor[gray]{0.9} \textbf{C3RL (Ours)} & \textbf{49.1\%}	&\underline{0.734}	&0.142 &\textbf{37.3\%}	&\textbf{0.660}	&0.246	&\textbf{64.0\%}	&0.680	&\underline{0.089} &\textbf{50.1\%}	&\underline{0.691}	&0.159\\
            \bottomrule[1pt]\midrule
            
        \end{tabular}%
    }
    \caption{Qwen2.5VL-7B-Instruct and Llama-3.2-3B-Instruct results across in-domain test set $D_{test}$ , Agieval, and MMLU benchmarks. Among RL methods, the best values are highlighted in bold, and the suboptimal values are underlined. Our proposed C3RL method is shaded. All the methods except for Self Consistency are tested three times with seed=0,42,2025 separately and the results are averaged. The maximum standard deviation among all results is $\pm$0.9\%, which is in a reasonable range.}
    \label{tab:calibration_text_combine}
\end{table*}

\begin{table*}[h]
    \centering
    \renewcommand{\arraystretch}{0.8}
    \setlength{\tabcolsep}{4pt}
    \resizebox{\textwidth}{!}{%
        \begin{tabular}{l|ccc|ccc|ccc|ccc}
            \toprule[1pt]\midrule
            ~ & \multicolumn{3}{c}{\textbf{MMMU}} 
              & \multicolumn{3}{c}{\textbf{MathVista}} 
              & \multicolumn{3}{c}{\textbf{LogicVista}} 
              & \multicolumn{3}{c}{\textbf{Avg.}} 
              \\
            \cmidrule(lr){2-4} \cmidrule(lr){5-7} \cmidrule(lr){8-10} \cmidrule(lr){11-13}
            \textbf{Method} & Acc$\uparrow$ & AUROC$\uparrow$ & ECE$\downarrow$ & Acc$\uparrow$ & AUROC$\uparrow$ & ECE$\downarrow$ & Acc$\uparrow$ & AUROC$\uparrow$ & ECE$\downarrow$ & Acc$\uparrow$ & AUROC$\uparrow$ & ECE$\downarrow$ \\
            \midrule
            \multicolumn{13}{c}{\textbf{Qwen2.5VL-7B-Instruct}} \\
            \midrule
            Base & 0.473 & 0.571 & 0.381 & 0.543 & 0.623 & 0.284 & 0.368 & 0.541 & 0.493 &0.461		&0.578	&0.386 \\
            SC & 0.509 & 0.621 & 0.211  & 0.591 & 0.818 & 0.077 & 0.446 & 0.678 & 0.157& 0.515		&0.706	&0.148 \\
            SFT+Ref & 0.368	&	0.651&	0.095	 &	0.452&		0.702	&0.091&	0.306	&	0.578&	0.092 &0.375		&0.644&	0.093
            \\
            \midrule
            RLVR & 0.520 & 0.525 & 0.417  & \underline{0.614} & 0.595 & 0.304 & \textbf{0.467} & 0.522 & 0.495  &\underline{0.534}		&0.547	&0.405\\
            SaySelf & 0.331 & \textbf{0.784} & 0.160  & 0.431 & \textbf{0.822} & \underline{0.094} & 0.261 & \textbf{0.739} & \textbf{0.152} &0.341		&\textbf{0.782}	&\underline{0.135}\\
            RLCR & 0.509 & 0.651 & \underline{0.146}  & 0.600 & \underline{0.690} & 0.117 & 0.431 & \underline{0.608} & 0.284 &0.513		&\underline{0.650}	&0.182\\
            C3RL w/o Ref & \underline{0.524} & 0.613 & 0.173 & \textbf{0.622} & 0.669 & 0.139 & \underline{0.462} & 0.591 & 0.268 &\textbf{0.536}		&0.624	&0.193\\
            \rowcolor[gray]{0.9} \textbf{C3RL (Ours)} & \textbf{0.530} & \underline{0.655} & \textbf{0.096}  & \textbf{0.622} & 0.682 & \textbf{0.074} & 0.446 & 0.596 & \underline{0.158} &0.533		&0.644	&\textbf{0.109}\\
            \bottomrule[1pt]\midrule
        \end{tabular}%
    }
    \caption{Qwen2.5VL-7B-Instruct results across multimodal reasoning benchmarks MMMU, MathVista and LogicVista. Among RL methods, the best values are highlighted in bold, and the suboptimal values are underlined. Our proposed C3RL method is shaded and achieves superior calibration performance with the lowest ECE scores on MMStar and MathVista. All the methods except for Self Consistency are tested three times with seed=0,42,2025 separately and the results are averaged. The maximum standard deviation among all results is $\pm$1.0\%, which is in a reasonable range.}
    \label{tab:calibration_multimodal}
\end{table*}

We use the following metrics to test LLMs' ability to give correct and calibrated responses:

\textbf{Accuracy ($\uparrow$): } To measure the correctness of LLMs on the evaluation set: $\mathrm{Acc} = {{1\over N} {\sum_{i=1}^{N}\mathbb{I}(a_{q_i})}}$

\textbf{AUROC ($\uparrow$): } Area Under ROC Curve measures the ability to distinguish between correct and incorrect answers under different confidence thresholds: 
\begin{equation}
    \mathrm{AUROC} = \int_1^{10} \mathrm{TPR}(\mathrm{FPR}(t)^{-1}) \, d (t), 
\end{equation}
where $t \in [1,10]$ is confidence threshold, TPR is true positive rate, and FPR is false positive rate.


\textbf{ECE ($\downarrow$): } We choose Binned Expected Calibration Error to describe the  gap between verbalized confidence and accuracy:
\begin{equation}
    \mathrm{ECE} = {m\over N} \sum_{i=1}^{\lceil {N\over m} \rceil} |{\mathrm{Acc}(B_i) - \mathrm{Conf}(B_i)|} , 
\end{equation}
where $N$ is the number of samples, $m=20$ is the fixed bin size , $B_i$ is the set of data in the $i$-th bin.

\subsubsection{Baselines}
We compare the following methods: 

\textbf{Non-RL Methods }(1) {Base}: We directly ask the untrained model to output both answer and associated confidence ; (2) {Self Consistency}({SC})~\citep{wang2023self-consistency} : Each question is sampled 10 times and the final answer is obtained from majority voting; (3) {Supervised Finetuning with reference accuracy}({SFT+Ref})~\cite{sftref}: The model is supervised fine-tuned to calibrate with reference accuracy calculated in 10-time sampling.

\textbf{RL Methods }(1) {RLVR}: The off-the-shelf method using the correctness reward ; (2) {SaySelf}~\citep{xu2024sayself} :The reward simply contains calibration Brier score; (3) {RLCR}~\citep{damani2025RLCR}: Based on SaySelf, RLCR adds the correctness score to ensure the performance, which is current state-of-the-art RL method to tradeoff between performance and calibration; (4) {C3RL (Ours) and variants}: C3RL integrates the correctness reward, calibration reward and reference accuracy reward together. We exclude the reference accuracy reward (C3RL w/o Ref) to conduct ablation study.


\subsubsection{Results: Calibration-Accuracy Trade-off}
Table~\ref{tab:calibration_text_combine} shows the comparison of different methods in both accuracy and calibration metrics.

\textbf{In-domain Performance }
As shown in Table~\ref{tab:calibration_text_combine}, all calibration-aware reward methods consistently outperform the basic RLVR approach in both AUROC and ECE on the in-domain test set $\mathcal{D}_{test}$. Notably, Qwen2.5VL-7B-Instruct trained after C3RL achieves highly competitive performance, closely matching RLVR in accuracy while attaining the similar superior ECE (0.076) as the best-calibrated method, SaySelf. This demonstrates C3RL's strong ability to maintain high accuracy alongside excellent calibration, even though it does not explicitly surpass the current state-of-the-art RLCR in performance. 

\begin{table*}[ht]
    \centering
    \renewcommand{\arraystretch}{0.8}
    \setlength{\tabcolsep}{4pt} 
    \resizebox{\textwidth}{!}{%
        \begin{tabular}{l|ccc|ccc|ccc|}
            \toprule[1pt]\midrule
            ~ & \multicolumn{3}{c}{\textbf{In-domain}} 
              & \multicolumn{3}{c}{\textbf{OOD (Text)}} 
              & \multicolumn{3}{c}{\textbf{OOD (Multimodal)}} \\
            \cmidrule(lr){2-4} \cmidrule(lr){5-7} \cmidrule(lr){8-10}
            \textbf{Method} & Avg. N$\downarrow$ &Saved Times$\uparrow$ & Acc$\uparrow$ & Avg. N$\downarrow$ & Saved Times$\uparrow$& Acc$\uparrow$ & Avg. N$\downarrow$ & Saved Times$\uparrow$ & Acc$\uparrow$ \\
            \midrule
            MV~\citep{wang2023self-consistency} &	64 & $\times$1 &	67.3\%	& 64 & $\times$1 &	78.1\% &	64 & $\times$1 &	57.8\% \\
            AC~\citep{aggarwal2023AC}&	13.71& $\times$4.67 &	67.4\% &	12.87& $\times$4.97 &	78.1\%&	15.24& $\times$4.20 &	57.9\% \\
            \textbf{CAS (Ours)} & \textbf{10.91} & \textbf{$\times$5.87} &	67.5\%	& \textbf{5.19} & \textbf{$\times$12.33}	&78.2\%&	\textbf{7.13} & \textbf{$\times$8.98} &	57.9\%  \\
            \bottomrule[1pt]\midrule
        \end{tabular}%
    }
    \caption{Comparative results of \textbf{MV} (Majority Voting), \textbf{AC} (Adaptive Consistency) and \textbf{CAS} (Confidence-based Adaptive Test Time Scaling) on In-domain, OOD (text), and OOD (multimodal). By adjusting the threshold $P_{\mathrm{stop}}$ of AC and CAS, we compare their needed inference budget Avg. N to match the accuracy of MV($N=64$).}
    \label{tab:test-time-scaling}
\end{table*}
\textbf{Generalization } 
As shown in Table~\ref{tab:calibration_text_combine} and Table~\ref{tab:calibration_multimodal}, C3RL exhibits remarkable generalization capabilities on out-of-domain(OOD) datasets. Compared to other top-calibrated RL methods, Qwen2.5VL-7B-Instruct trained after C3RL is superior to SaySelf—the best RL strategy in calibration—in ECE metric and surpasses it greatly in accuracy (7.9\% in text sets and 19.2\% in multimodal sets). Moreover, C3RL consistently outperforms the current state-of-the-art RLCR across both accuracy and calibration metrics on OOD evaluations, demonstrating an optimal and robust balance between performance and reliability. 
When generalized to a different family model Llama-3.2-3B-Instruct~\cite{llama3_2_3b}, although achieving better calibration, RLCR sacrifices more performance and can’t match RLVR in accuracy, which means there are more deliberately low-confidence incorrect answers. On the contrary, C3RL can still surpass RLVR in performance and decrease calibration error greatly at the same time, proving the stability of C3RL to different model families and sizes.

\textbf{Ablation Study }
When the reference accuracy reward is ablated (C3RL w/o Ref), the model remains competitive accuracy but suffers a consistent degradation in calibration. This clearly displays the specific contribution of the reference accuracy reward to enhancing calibration capability.

\subsubsection{Analysis: C3RL Advantage}
The advantage of C3RL over RLCR in balancing performance and calibration is due to the following two reasons:
(1) Reference accuracy information: The ablation study has demonstrated the effectiveness of reference accuracy reward in advancing calibration;
(2) Importance of learning confidence threshold: RLCR, driven by the Brier Score (see \ref{appendix: reward details}), tends to train LLMs to indicate totally confident ($c$=10) or totally inconfident ($c$=1) responses, which degrades to a binary case. On the contrary, C3RL guides the model to learn the confidence threshold $t$=5 and gives fine-grained confidence to distinguish between certainty and uncertainty. The difference in confidence distribution between C3RL and RLCR is demonstrated in~\ref{appendix: confidence distribution}.

\subsection{Can CAS Improve Test Time Scaling?}
\label{sec: CAS}
Given the impressive results of self-consistency in Table~\ref{tab:calibration_text_combine}, it is natural to combine verbalized confidence and self-consistency together.
Although there have been test time scaling methods leveraging confidence to enhance accuracy (e.g. confidence-weighted majority voting), they fail to apply calibrated confidence to saving inference budget. We conduct experiments to demonstrate the inference efficiency of Confidence-based Adaptive Test Time Scaling (CAS).

\subsubsection{Experimental Setup for CAS}
To evaluate the performance and inference efficiency of CAS, we employ the following settings: 

\textbf{Benchmarks }
\label{sec:CAS benchmarks}
We select 6 benchmarks of reasonable sizes (less than 2000 samples): (1) \textbf{Text}: MMLU validation set ~\cite{mmlu}, FOLIO~\citep{folio} and GSM8K~\citep{gsm8k}; (2) \textbf{Multimodal} : 
The same as CAS evaluation multimodal sets.
The details of the benchmarks are in~\ref{appendix:benchmark}.

\textbf{Baselines }

(1) 
\textbf{Majority Voting (MV)}~\citep{wang2023self-consistency}: Each question is generated $N$ times and the most frequent answer is voted.
(2) \textbf{Adaptive-Consistency (AC)}~\citep{aggarwal2023AC}: A strategy dynamically adjusts the number of samples based on a stopping criteria to detect dominant answers. The final answer is voted from the truncated samples. 


\textbf{Settings }
To evaluate all the methods equally, we compare the inference budget needed to achieve the same accuracy as Majority Voting($N$=64). It's guaranteed by controlling threhold $P_{\mathrm{stop}}$ for AC and CAS separately. We use the Qwen2.5VL-2.5-7B-Instruct model trained after C3RL.

\subsubsection{Results}

As shown in Table~\ref{tab:test-time-scaling}, our proposed CAS method demonstrates remarkable efficiency gains while maintaining competitive accuracy across all evaluation settings.

In terms of sampling efficiency, CAS achieves outstanding cost reduction compared to baseline methods.
Notably, CAS consistently outperforms the strongest baseline Adaptive-Consistency. The performance gap is particularly striking on OOD (Text) tasks, where CAS uses less than half of the sampling budget of Adaptive-Consistency (5.19 vs. 12.87 samples).

\begin{figure}[h]
    \centering
    \includegraphics[clip, width=0.9\columnwidth]{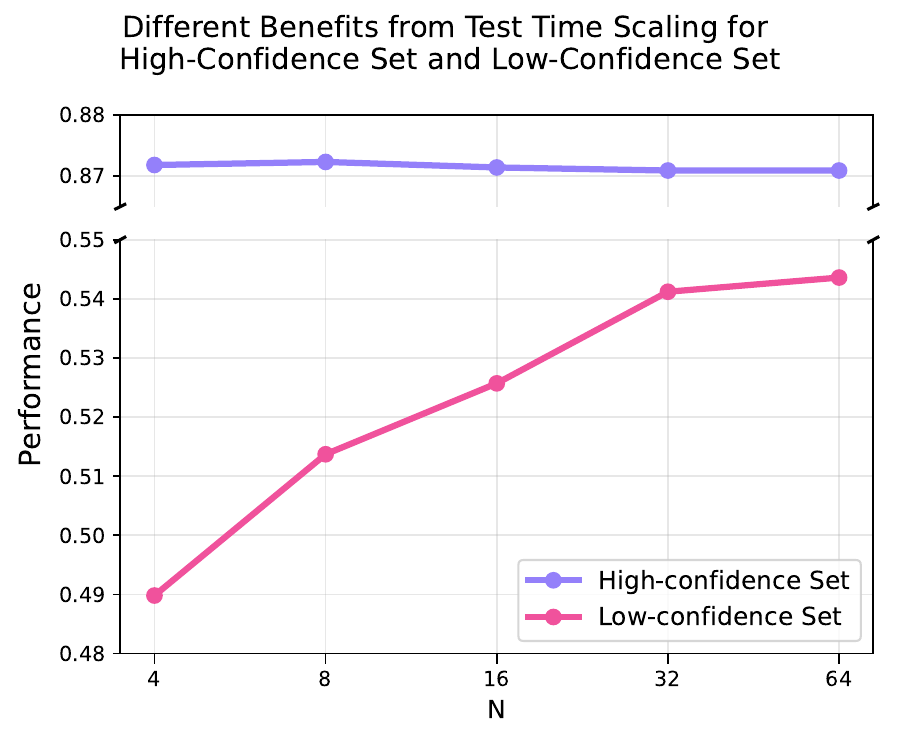}
    \caption{Different benefits from test time scaling for High-confidence set (CAS stopping $n\leq 4)$ and Low-confidence set (CAS stopping $n>4$)  on OOD (Text). The 0.55-0.86 range of y-axis is cut for clarity. The High-Confidence set benefits much less from test time scaling than the Low-Confidence set.}
    \label{fig:Benefits}
\end{figure}
\begin{figure*}[ht]
    \centering
    \includegraphics[trim=10 10 10 10, clip, width=1.9\columnwidth]{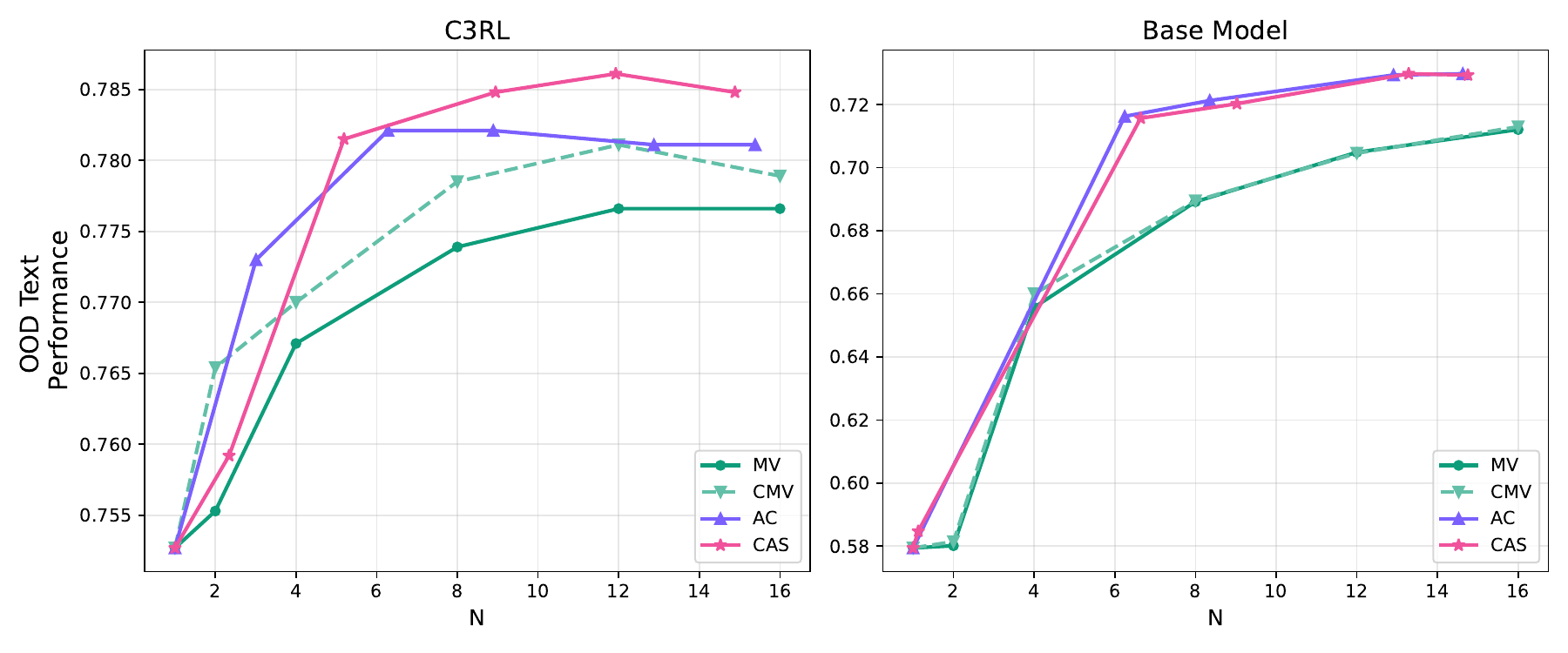}
    \caption{Performance of C3RL model vs. Base model with different test time scaling methods on OOD (Text). \textbf{MV} denotes Majority Voting, \textbf{CMV} denotes Confidence-weighted Majority Voting, \textbf{AC} denotes Adaptive Consistency, \textbf{CAS} denotes Confidence-based Adaptive Test Time Scaling. C3RL model shows performance stratification on different methods, while the base model improves little for confidence-based methods.}
    \label{fig:CAS}
\end{figure*}
These efficiency gains come with no cost of accuracy. 
This demonstrates that CAS can precisely identify high-quality reasoning paths early in the generation process, eliminating unnecessary computations without sacrificing final performance.

\subsubsection{Analysis: Imbalanced Budget Allocation}
CAS adjusts the inference budget from high-confidence sampling traces to low-confidence traces with early-stopping strategy. To verify the rationality of this imbalanced mechanism, we compare the benefits from test time scaling for high-confidence and low-confidence samples.

We use the C3RL model for generation and choose the three text benchmarks in Sec \ref{sec:CAS benchmarks} for evaluation. We divide the samples into two sets: High-confidence Set and Low-confidence Set.
For each question, if the CAS sampling time $n \leq 4$ (which means CAS stops early because the final answer reaches high confidence quickly), we categorize this question to High-confidence Set. Otherwise, we assign the sample to Low-confidence Set. We study how these two sets benefit differently from majority voting.

As shown in Figure \ref{fig:Benefits}, when test time $N$ increases to 64, the High-confidence Set remains unchanged in performance (<0.2\%), while the Low-confidence Set jumps dramatically from 49.0\% to 54.5\%. The fact that the Low-confidence Set benefits much more from test time scaling than its counterpart rationalizes the CAS strategy of shifting limited computing resource from high-confidence samples to low-confidence samples.

\subsection{Can C3RL contribute to CAS?}
\label{sec:C3RL-CAS}
To explore the influence of calibrated confidence in test time scaling, we compare two models: The base model and the model trained with C3RL. The Confidence-weighted Majority Voting (CMV) method is also evaluated to display the contribution of calibrated confidence.

As shown in Figure~\ref{fig:CAS}, C3RL+CAS outperforms all baseline methods when number of samples $N$ is sufficiently large ($\geq$8), demonstrating the scalable effectiveness of our approach.  A key observation is the stratification of test-time scaling strategies in the C3RL-trained model. In comparison, the base model benefits little from CAS than AC. This stark contrast underscores that confidence calibration through C3RL is prerequisite for CAS to unlock its full potential. The performance leap from base+CAS to C3RL+CAS demonstrates their synergistic relationship: C3RL provides well-calibrated verbalized confidence, and CAS efficiently leverages it for test-time scaling.

\section{Conclusion}
This work presents Correctness and Confidence Calibration Reinforcement Learning (C3RL) to support downstream Confidence-based Adaptive test time Scaling (CAS). By integrating correctness, calibration and prior reference information together, C3RL achieves the best trade-off between performance and calibration, providing credible verbalized confidence. Based on well-calibrated confidence, CAS takes inference most efficiently among test time scaling methods by adjusting computing resources to low-confidence samples. Our results highlight a unified pipeline for reliable LLMs and efficient inference under limited resources.

\section*{Limitations}
There are several limitations for our C3RL-CAS framework. First, the high-quality training data in science domain is not sufficient compared to Maths and logic reasoning domains, which may result in training bias. Second, although C3RL achieves best accuracy-calibration trade-off, its calibration ability is still inferior to SaySelf, which means if we want to enhance the model accuracy, we have to sacrifice calibration. Third, the advantage of CAS over Adaptive-Consistency in small generation cases ($N$<8) is not considerable, which requires more reasonable stopping criteria and better-calibrated confidence from C3RL.



\bibliography{latex/bibliography}

\clearpage
\appendix

\section{Appendix}
\label{sec:appendix}
\subsection{Experimental Data}
\subsubsection{Training Dataset Construction}
\label{appendix:training set}
We modify the two reasoning datasets Webinstruct-verified~\cite{webinstruct} and Numina-NLI~\cite{numina_math} to support regex check of response correctness.

The Webinstruct-verified dataset is modified from the original 231,833 samples. We filter the science data (domains in "physics", "chemistry", "biology", "computer science", "engineering") and select the Integer answer samples. The selected sample number is 9,246. 

For Numina-NLI and Webinstruct-verified datasets whose answers are numerical expressions, we check whether the generated samples are "math valid". We exclude the questions with no valid responses among 10 generations. For LogicNLI~\cite{logicnli} and LogiQA~\cite{logiqa} datasets, we follow the selection process specified in C2RM~\cite{c2rm}. 

Finally, $\mathcal{D}$ composes of the following training datasets: Numina-NLI (69,043), Webinstruct-verified (4,488), LogicNLI (16,000) and LogiQA (17,730).

\subsubsection{Benchmark Details}
\label{appendix:benchmark}
We evaluate reasoning and logic capabilities across reasoning, math and logic benchmarks with text and multimodal datasets:

\textbf{Text:} (1) AGIEval~\cite{agieval}: 8269 problems from college admission tests, law school admission tests, math competitions, lawyer qualification tests, and national civil service exams. We use the combination of train, validation and test set for evaluation; (2) OlympiadBench (text)~\cite{olympiadbench}: Olympiad-level questions including problems from maths and physics competitions. We select the 2,673 text samples for evaluation; (3) MMLU~\cite{mmlu}: A multiple choice benchmark across STEM, the humanities and the social science. The validation set contains 1,531 samples and the test set includes 14,041 samples. For C3RL evaluation, we use valiation set + test set; For CAS evaluation, we only use the validation set; (4) Folio: 223 multiple-choice questions with logical reasoning tasks; (5) GSM8K (Grade School Math 8K)~\cite{gsm8k}: 8.5K high quality linguistically diverse grade school math word problems, and we only use the 1,319 questions in test set.

\textbf{Multimodal } 
(1) MathVision~\cite{mathvision}: High-quality maths problems with visual contexts from real math competitions; the test set contains 3,040 samples; (2) MathVista~\cite{mathvista}: 1,000 challenging Maths reasoning questions with visual contexts; (3) LogicVista~\cite{logicvista}: 448 multimodal questions focusing on logical reasoning tasks (spatial, deductive, inductive, numeric, and mechanical reasoning); (4) MMStar~\cite{mmstar}: 1,500 human-curated challenging samples to test the multimodal abilities in general tasks like fine-grained perception, counting and logic reasoning.

\subsection{Training Details}
This section complements the details to the training process.
\subsubsection{Generation Prompt}
\begin{table}[h]
\small
\centering
\scalebox{1}{
\begin{tcolorbox}[colback=gray!01,
                  width=\columnwidth,
                  arc=1mm, auto outer arc,
                  breakable,
                  left=0.9mm, right=0.9mm,
                  boxrule=0.9pt, colbacktitle = black!65!black,
                  title={Prompt to generate answers}
                 ]
    "Answer the question following this exact format: **Strict Requirements**: **Structure**: "
    "After reasoning, output ONLY the [final answer] within $\backslash$boxed\{\{\}\} format: <think> [your analysis] </think>\textbackslash n "
    "<ANSWER is: $\backslash$boxed\{Your answer\}> **No Extra Text**: Do not explain your answer outside<think> tags.
    Put your answer into <ANSWER is: $\backslash$boxed\{Your answer\}>. Based on your answer, please attach a confidence signal ranging from 1-10 to specify whether you are certain about your answer. 1 means you are totally uncertain (strong inconfidence), while 10 means you are totally certain (strong confidence). If you need more information to answer the question, please attach 1. We will compare your answer with the ground truth to check the correctness. If your answer is correct and accompanied by strong confidence, you will be rewarded; if your answer is incorrect but assigned strong confidence, you will be punished. The signal should be in the format of <CONFIDENCE:NUMBER>, where NUMBER ranges from 1 to 10, directly appended to your answer. The last line of your output should be in the format: <ANSWER is: $\backslash$boxed\{Your answer\}>|<CONFIDENCE:NUMBER>."

\end{tcolorbox}
}
\caption{Prompt to generate answers and elicit verbalized confidence}
\label{tab:prompt_answer}
\end{table}
\label{appendix: answer generation}
In Sec \ref{training dataset}, we need to generate 10 samples per question. 
The model answer generation prompt is shown in Table~\ref{tab:prompt_answer}. The prompt: (1) restricts the formats of responses so that the answers can be extracted by rule and (2) elicits the verbalized confidence with an elaborated description of certainty and uncertainty.

\begin{figure*}[htbp]
    \centering
    \begin{subfigure}[b]{0.45\textwidth}
        \centering
        \includegraphics[width=\textwidth]{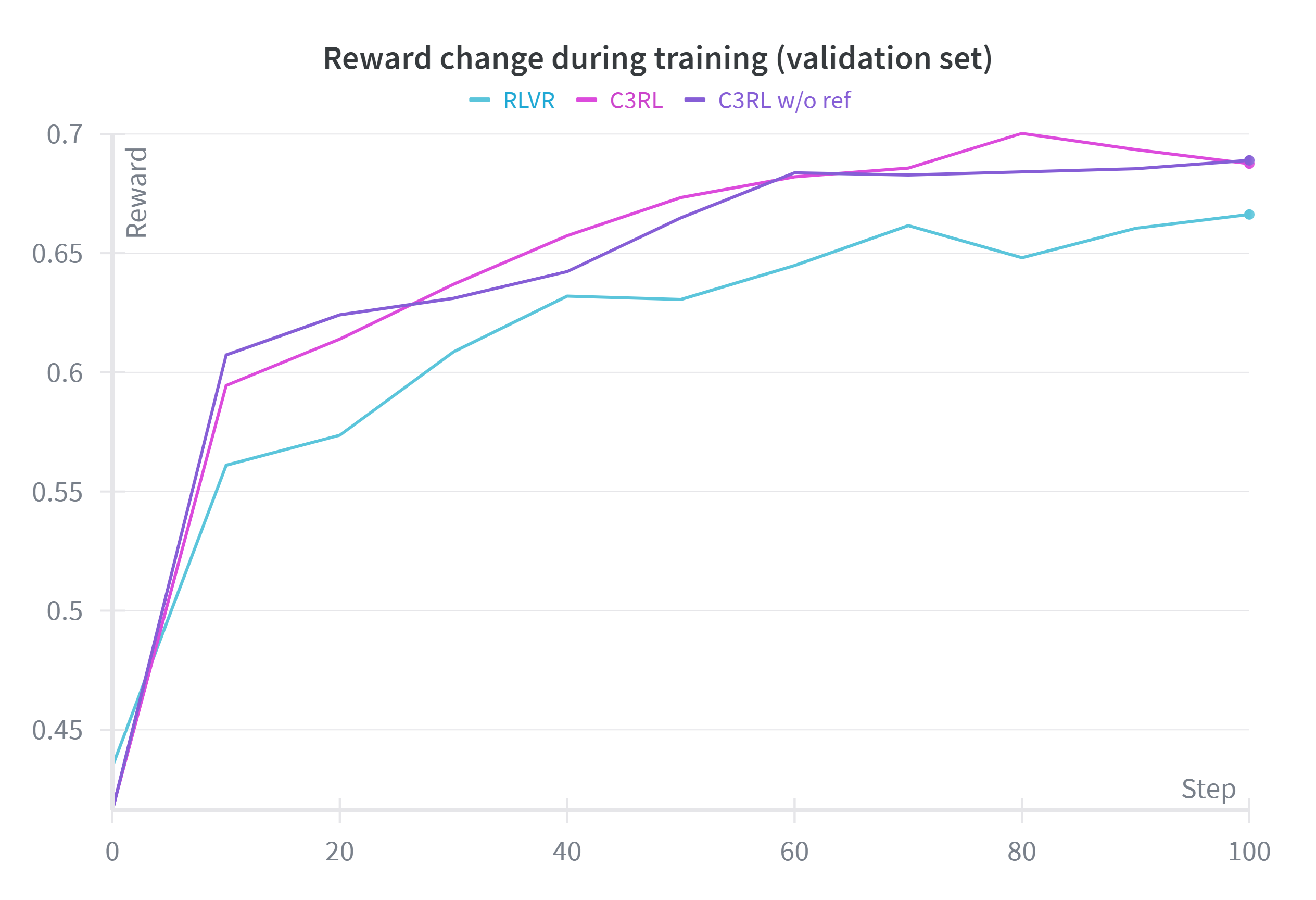} 
        \caption{RLVR, C3RL and C3RL w/o ref rewards} 
        \label{fig:sub1}
    \end{subfigure}
    \hfill 
    \begin{subfigure}[b]{0.45\textwidth}
        \centering
        \includegraphics[width=\textwidth]{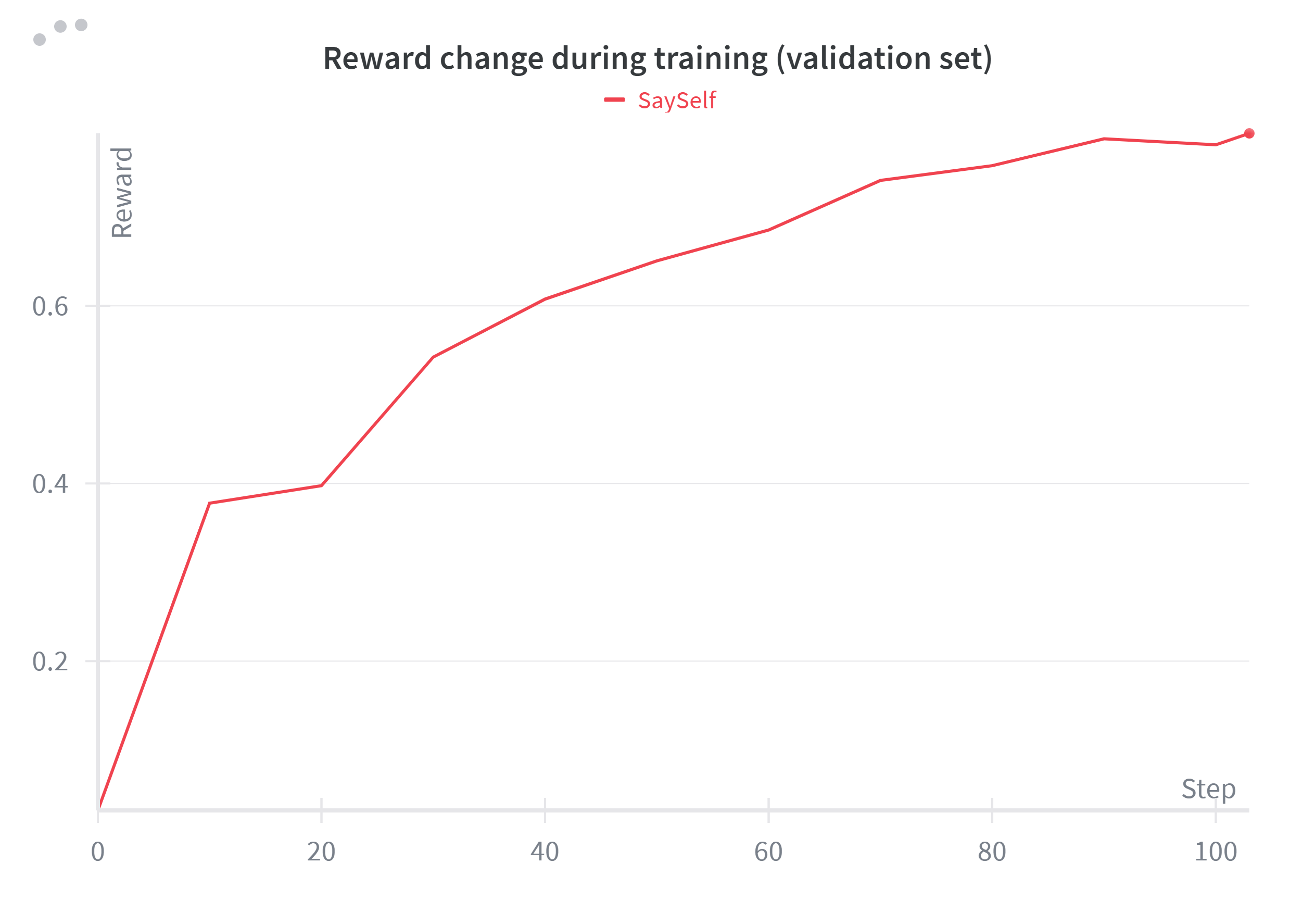}
        \caption{SaySelf reward}
        \label{fig:sub2}
    \end{subfigure}
    
    \begin{subfigure}[b]{0.45\textwidth}
        \centering
        \includegraphics[width=\textwidth]{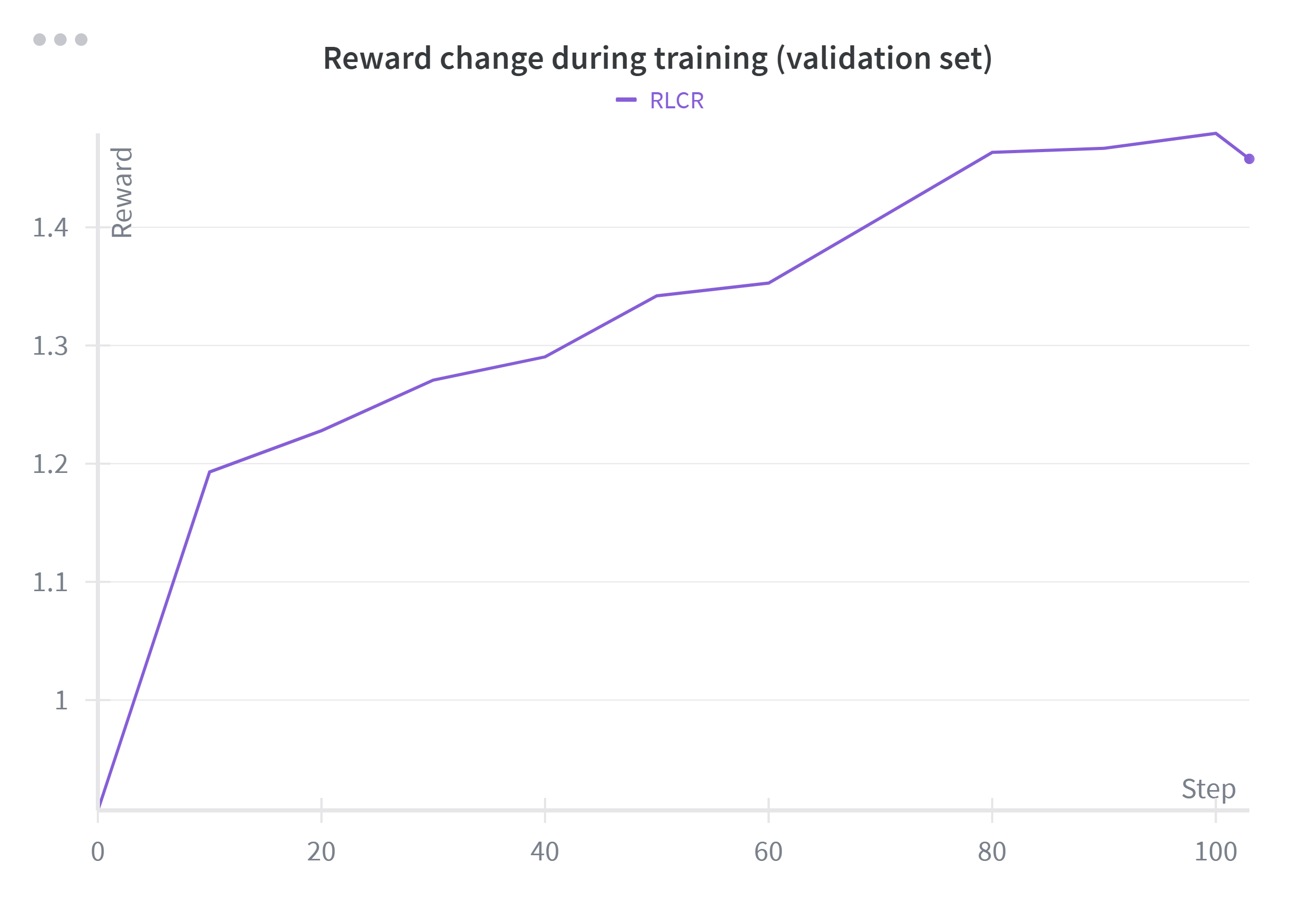}
        \caption{RLCR reward}
        \label{fig:sub3}
    \end{subfigure}
    \hfill
    \begin{subfigure}[b]{0.45\textwidth}
        \centering
        \includegraphics[width=\textwidth]{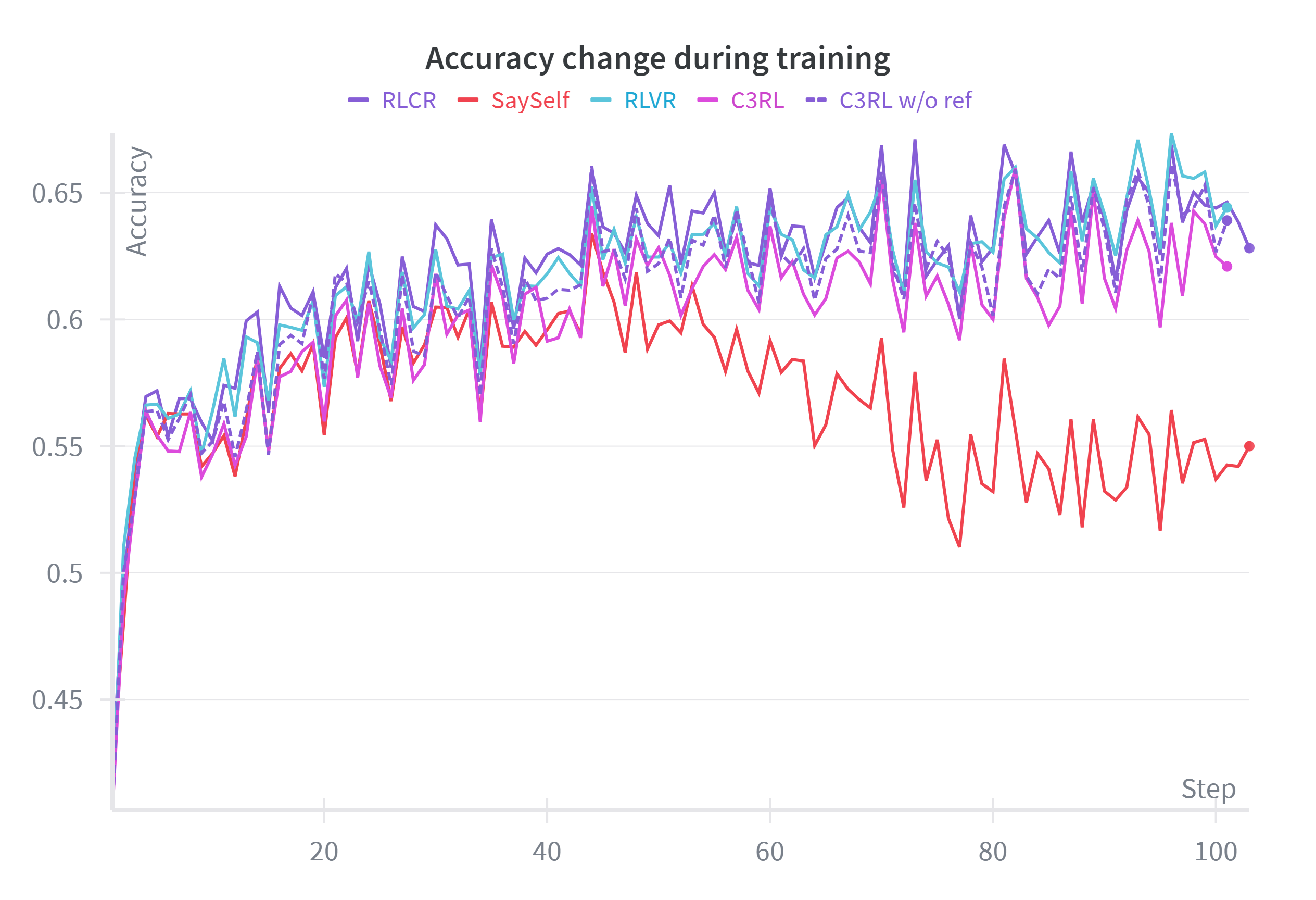}
        \caption{Accuracy comparison during training}
        \label{fig:sub4}
    \end{subfigure}
    
    \caption{Training curves} 
    \label{fig:training_figure}
\end{figure*}

\subsubsection{Training Settings}
\label{appendix:training setting}
We set the generation hyperparameters as shown in Table~\ref{tab:hyperparameters}. We employ VeRL~\cite{verl} framework and use 8 NVIDIA-A800-SXM4-80GB GPUs to support C3RL training of Qwen2.5-VL-7B-Instruct.
\begin{table}[htbp]
\centering
\begin{tabular}{lc}
\toprule
\textbf{Config} & \textbf{Value} \\
\midrule
actor-lr & $1 \times 10^{-6}$ \\
kl\_coef & 0.001 \\
max\_prompt\_length & 2,048 \\
max\_response\_length & 2,048 \\
train\_batch\_size & 1,024 \\
ppo\_mini\_batch\_size & 256 \\
clip\_ratio & 0.20 \\
sample\_temperature & 0.7 \\
rollout.n & 10 \\
total\_training\_steps & 103 \\
\bottomrule
\end{tabular}
\caption{Training hyper-parameters during C3RL}
\label{tab:hyperparameters}
\end{table}

\subsubsection{Reward Details}
\label{appendix: reward details}
For all the RL in the experiments, we first assign a format reward $R_\mathrm{{format}}(a_q)$ to the responses:  
Following the prompt designed in Table~\ref{tab:prompt_answer}, if both answer and confidence can be extracted, we give a format reward $R_\mathrm{{format}}(a_q)$=+0.1; otherwise, the final reward $R_\mathrm{{C3RL}}$ will be set to -1 directly.

For C3RL reward, $R_\mathrm{{correctness}}(a_q)$ = 0.9$\mathbb{I}(a_q=a^*)$ so that a correct answer will have $R_\mathrm{{correctness}}(a_q)$ + $R_\mathrm{{format}}(a_q)$ = 1 (equal to RLVR reward). 

For SaySelf reward, we use the Brier Score $R_\mathrm{SaySelf}$=1-2$(\mathbb{I}(a_q=a^*)-\mathrm{conf}(r)/10)^2$. 

For RLCR reward, we increment a correctness reward based on SaySelf reward: $R_\mathrm{RLCR}=\mathbb{I}(a_q=a^*)-(\mathbb{I}(a_q=a^*)-\mathrm{conf}(r)/10)^2$.

\subsubsection{Training Reward Curves}
The reward curves and the performance curves are shown in Figure~\ref{fig:training_figure}. All the methods converge after 100 steps. Figure~\ref{fig:sub1} displays the effectiveness of calibration reward (C3RL w/o Ref curve is higher than RLVR curve) and reference accuracy reward (C3RL curve is higher than C3RL w/o Ref curve). As for performance, ~\ref{fig:sub4} shows that all the strategies match at the similar level except for SaySelf, which demonstrates the significance of correctness reward to guarantee accuracy.

\subsection{Evaluation Details}
 \begin{table*}[ht]
    \centering
    \renewcommand{\arraystretch}{1.1}
    \setlength{\tabcolsep}{4pt}
    \resizebox{\textwidth}{!}{%
        \begin{tabular}{l|ccc|ccc|ccc|ccc}
            \toprule[1pt]\midrule
                \textbf{Text}
              & \multicolumn{3}{c}{\textbf{In-domain}} 
              & \multicolumn{3}{c}{\textbf{Agieval}} 
              & \multicolumn{3}{c}{\textbf{Olympiad}} 
              & \multicolumn{3}{c}{\textbf{MMLU}}  \\
            \cmidrule(lr){2-4} \cmidrule(lr){5-7} \cmidrule(lr){8-10} \cmidrule(lr){11-13}
            \textbf{Method} & Acc$\uparrow$ & AUROC$\uparrow$ & ECE$\downarrow$ & Acc$\uparrow$ & AUROC$\uparrow$ & ECE$\downarrow$ & Acc$\uparrow$ & AUROC$\uparrow$ & ECE$\downarrow$ & Acc$\uparrow$ & AUROC$\uparrow$ & ECE$\downarrow$  \\
            \midrule
            \multicolumn{13}{c}{\textbf{Qwen2.5VL-7B-Instruct}} \\
            \midrule
            Base & 45.3\% & 0.531 & 0.448 & 51.0\% & 0.543 & 0.375 & 20.3\% & 0.594 & 0.709 & 68.3\% & 0.587 & 0.188 \\
            SC & 53.9\% & 0.762 & 0.088 & 53.7\% & 0.711 & 0.208 & 26.6\% & 0.783 & 0.090 & 69.9\% & 0.759 & 0.141 \\
            SFT+Ref & 48.0\% & 0.720 &	0.098 & 49.1\%	&0.753&	0.105&	10.3\%	&0.732	&0.141	&62.1\%	&0.706	&0.098
            \\ 
            \midrule
            RLVR & \textbf{66.6\%} & 0.514 & 0.285 &  \textbf{58.1\%} & 0.534 & 0.377 & \textbf{37.6\%} & 0.514 & 0.616 & 71.1\% & 0.602 & 0.239 \\
            SaySelf & 55.0\% & \textbf{0.807} & \textbf{0.075} & 50.8\% & \textbf{0.745} & \textbf{0.147} & 28.3\% & \underline{0.799} & \textbf{0.071} & 64.0\% & \textbf{0.731} & 0.129 \\
            RLCR & \underline{65.1}\% & \underline{0.751} & 0.109 & 57.5\% & 0.664 & 0.205 & 35.4\% & \textbf{0.805} & \underline{0.147} & 70.7\% & 0.607 & 0.175 \\
            C3RL w/o Ref & 64.9\% & 0.700 & 0.119 & \textbf{58.1\%} & 0.682 & 0.197 & 35.9\% & 0.745 & 0.263 & \underline{71.2\%} & 0.648 & \underline{0.111}  \\
            \rowcolor[gray]{0.9} \textbf{C3RL (Ours)} & 64.4\% & 0.748 & \underline{0.076} & \underline{57.8\%} & \underline{0.688} & \underline{0.154} & \underline{37.5\%} & 0.783 & 0.191 & \textbf{71.4\%} & \underline{0.653} & \textbf{0.092}  \\
            \bottomrule[1pt]
            \multicolumn{13}{c}{\textbf{Llama-3.2-3B-Instruct}} \\
            \midrule
            Base & 37.2\%	& 0.563	&0.134 & 34.8\%	&	0.557	&0.423	&8.3\%		&0.559	&0.379	&59.6\%		&0.586	&0.271 \\
            SC &  40.8\%	&	0.787&	0.091& 28.6\%	&	0.644	&0.352&	8.6\%	&	0.631	&0.120&	59.9\%	&	0.733	&0.158 \\
            SFT+Ref & 37.6\%	&0.572	&0.168 &32.1\%	&0.594	&0.240	&5.2\%		&0.583	&0.170	&52.8\%	&0.628	&0.162 \\
            \midrule
            RLVR & 45.9\%	&0.587	&0.463 &\textbf{37.3\%}	&0.522	&0.468	&\textbf{17.0\%}	&0.593	&0.754	&63.4\%	&0.537	&0.208 \\
            SaySelf & 13.1\%	&0.676	&\textbf{0.060} &19.4\%		&0.611	&\textbf{0.072}	&3.3\%	&0.654	&\underline{0.083}	&37.4\%	&\textbf{0.715}	&0.110 \\
            RLCR & 43.3\%	&\textbf{0.799}	&\underline{0.072} &34.4\%	&0.658	&\underline{0.111}	&12.8\%	&\textbf{0.657}	&\textbf{0.077}	&62.3\%	&0.670	&0.153 \\
            C3RL w/o Ref & \underline{47.3\%}	&0.697	&0.136 &\underline{37.2\%}	&0.652	&0.218	&\underline{16.2\%}	&\underline{0.656}	&0.278	&\underline{63.5\%}	&\underline{0.688}	&\textbf{0.087} \\
            \rowcolor[gray]{0.9} \textbf{C3RL (Ours)} & \textbf{49.1\%}	&\underline{0.734}	&0.142 &\textbf{37.3\%}	&\textbf{0.660}	&0.246	&15.8\%	&0.641	&0.289	&\textbf{64.0\%}	&0.68	&\underline{0.089} \\
            \bottomrule[1pt]\midrule
            
        \end{tabular}%
    }
    \caption{Qwen2.5VL-7B-Instruct and Llama-3.2-3B Instruct results across in-domain test set $D_{test}$ , Agieval, Olympiad, and MMLU benchmarks. Among RL methods, the best values are highlighted in bold, and the suboptimal values are underlined. Our proposed C3RL method is shaded. All the methods except for Self Consistency are tested three times with seed=0,42,2025 separately and the results are averaged. The maximum standard deviation among all results is 0.9\%, which is in a reasonable range.}
    \label{tab:calibration_text_combine_original}
\end{table*}
\begin{table*}[h]
    \centering
    \renewcommand{\arraystretch}{0.9}
    \setlength{\tabcolsep}{4pt}
    \resizebox{\textwidth}{!}{%
        \begin{tabular}{l|ccc|ccc|ccc|ccc}
            \toprule[1pt]\midrule
            ~ & \multicolumn{3}{c}{\textbf{MMMU}} 
              & \multicolumn{3}{c}{\textbf{MathVision}} 
              & \multicolumn{3}{c}{\textbf{MathVista}} 
              & \multicolumn{3}{c}{\textbf{LogicVista}} \\
            \cmidrule(lr){2-4} \cmidrule(lr){5-7} \cmidrule(lr){8-10} \cmidrule(lr){11-13}
            \textbf{Method} & Acc$\uparrow$ & AUROC$\uparrow$ & ECE$\downarrow$ & Acc$\uparrow$ & AUROC$\uparrow$ & ECE$\downarrow$ & Acc$\uparrow$ & AUROC$\uparrow$ & ECE$\downarrow$ & Acc$\uparrow$ & AUROC$\uparrow$ & ECE$\downarrow$ \\
            \midrule
            \multicolumn{13}{c}{\textbf{Qwen2.5VL-7B-Instruct}} \\
            \midrule
            Base & 0.473 & 0.571 & 0.381 & 0.242 & 0.508 & 0.680 & 0.543 & 0.623 & 0.284 & 0.368 & 0.541 & 0.493 \\
            SC & 0.509 & 0.621 & 0.211 & 0.282 & 0.700 & 0.193 & 0.591 & 0.818 & 0.077 & 0.446 & 0.678 & 0.157 \\
            SFT+Ref & 0.368	&	0.651&	0.095&	0.181	&	0.613	&0.083 &	0.452&		0.702	&0.091&	0.306	&	0.578&	0.092
            \\
            \midrule
            RLVR & 0.520 & 0.525 & 0.417 & 0.246 & 0.512 & 0.738 & \underline{0.614} & 0.595 & 0.304 & \textbf{0.467} & 0.522 & 0.495 \\
            SaySelf & 0.331 & \textbf{0.784} & 0.160 & 0.184 & \textbf{0.694} & \textbf{0.133} & 0.431 & \textbf{0.822} & \underline{0.094} & 0.261 & \textbf{0.739} & \textbf{0.152} \\
            RLCR & 0.509 & 0.651 & \underline{0.146} & 0.254 & 0.626 & \underline{0.253} & 0.600 & \underline{0.690} & 0.117 & 0.431 & \underline{0.608} & 0.284 \\
            C3RL w/o Ref & \underline{0.524} & 0.613 & 0.173 & \underline{0.273} & 0.607 & 0.382 & \textbf{0.622} & 0.669 & 0.139 & \underline{0.462} & 0.591 & 0.268 \\
            \rowcolor[gray]{0.9} \textbf{C3RL (Ours)} & \textbf{0.530} & \underline{0.655} & \textbf{0.096} & \textbf{0.279} & \underline{0.627} & 0.290 & \textbf{0.622} & 0.682 & \textbf{0.074} & 0.446 & 0.596 & \underline{0.158} \\
            \bottomrule[1pt]\midrule
        \end{tabular}%
    }
    \caption{Qwen2.5VL-7B-Instruct results across multimodal reasoning benchmarks MMMU, MathVision, MathVista and LogicVista. Among RL methods, the best values are highlighted in bold, and the suboptimal values are underlined. Our proposed C3RL method is shaded and achieves superior calibration performance with the lowest ECE scores on MMStar and MathVista. All the methods except for Self Consistency are tested three times with seed=0,42,2025 separately and the results are averaged. The maximum standard deviation among all results is 1.0\%, which is in a reasonable range.}
    \label{tab:calibration_multimodal_original}
\end{table*}

\subsubsection{Evaluation Settings}
\label{appendix:evaluation settings}
The evaluation hyperparameters keeps exactly the same as the generation hyperparameters in training process~\ref{appendix:training setting}. The evaluation prompts also follow the confidence elicitation prompts in Table~\ref{tab:prompt_answer}. We set the decoding temperature 0.7, top\_k=-1, top\_p=1, and seed=0, 42, 2025 separately for parallel experiments.

\subsubsection{Detailed Evaluation Results}
\label{appendix: detailed C3RL}
The C3RL evaluation of Qwen2.5VL-7B-Instruct~\cite{qwen2.5-vl} for all the datasets are elaborated in Table~\ref{tab:calibration_text_combine_original} and Table~\ref{tab:calibration_multimodal_original}. A very interesting phenomenon is that on all the multimodal benchmarks and MMLU, RLVR shows worse calibration metrics than the base model, which echoes the assumption that RLVR can produce overconfident and poorly calibrated models. Moreover, our proposed C3RL algorithm has more outstanding ability on multimodal tasks than on text tasks, which demonstrates that its advantage over other RL methods can be enhanced when generalized to multimodal sets, suggesting the occurrence of general confidence awareness. 

\subsubsection{Results for Llama-3.2-3B-Instruct}
\label{appendix: llama results}
The evaluation of C3RL on Llama-3.2-3B-Instruct~\cite{llama3_2_3b} for text datasets are detailed in Table~\ref{tab:llama_calibration}. Although achieving better calibration, RLCR sacrifices more performance and can't match RLVR in accuracy, which means there are more deliberately low-confidence incorrect answers. On the contrary, C3RL can still surpass RLVR in performance and decrease calibration error greatly at the same time, proving the stability of C3RL to different model families and sizes.

\begin{table*}[ht]
    \centering
    \renewcommand{\arraystretch}{0.9}
    \setlength{\tabcolsep}{4pt}
    \resizebox{\textwidth}{!}{%
        \begin{tabular}{l|ccc|ccc|ccc|ccc}
            \toprule[1pt]\midrule
            ~ & \multicolumn{3}{c}{\textbf{In-domain}} 
                & \multicolumn{3}{c}{\textbf{Agieval}} 
              & \multicolumn{3}{c}{\textbf{Olympiad}} 
              & \multicolumn{3}{c}{\textbf{MMLU}} \\
            \cmidrule(lr){2-4} \cmidrule(lr){5-7} \cmidrule(lr){8-10} \cmidrule(lr){11-13}
            \textbf{Method} & Acc$\uparrow$ & AUROC$\uparrow$ & ECE$\downarrow$ & Acc$\uparrow$ & AUROC$\uparrow$ & ECE$\downarrow$ & Acc$\uparrow$ & AUROC$\uparrow$ & ECE$\downarrow$ & Acc$\uparrow$ & AUROC$\uparrow$ & ECE$\downarrow$ \\
            \midrule
            Base & 37.2\%	& 0.563	&0.134 & 34.8\%	&	0.557	&0.423	&8.3\%		&0.559	&0.379	&59.6\%		&0.586	&0.271 \\
            SC &  40.8\%	&	0.787&	0.091& 28.6\%	&	0.644	&0.352&	8.6\%	&	0.631	&0.120&	59.9\%	&	0.733	&0.158 \\
            SFT+Ref & 37.6\%	&0.572	&0.168 &32.1\%	&0.594	&0.240	&5.2\%		&0.583	&0.170	&52.8\%	&0.628	&0.162 \\
            \midrule
            RLVR & 45.9\%	&0.587	&0.463 &\textbf{37.3\%}	&0.522	&0.468	&\textbf{17.0\%}	&0.593	&0.754	&63.4\%	&0.537	&0.208 \\
            SaySelf & 13.1\%	&0.676	&\textbf{0.060} &19.4\%		&0.611	&\textbf{0.072}	&3.3\%	&0.654	&\underline{0.083}	&37.4\%	&\textbf{0.715}	&0.110 \\
            RLCR & 43.3\%	&\textbf{0.799}	&\underline{0.072} &34.4\%	&0.658	&\underline{0.111}	&12.8\%	&\textbf{0.657}	&\textbf{0.077}	&62.3\%	&0.670	&0.153 \\
            C3RL w/o Ref & \underline{47.3\%}	&0.697	&0.136 &\underline{37.2\%}	&0.652	&0.218	&\underline{16.2\%}	&\underline{0.656}	&0.278	&\underline{63.5\%}	&\underline{0.688}	&\textbf{0.087} \\
            \rowcolor[gray]{0.9} \textbf{C3RL (Ours)} & \textbf{49.1\%}	&\underline{0.734}	&0.142 &\textbf{37.3\%}	&\textbf{0.660}	&0.246	&15.8\%	&0.641	&0.289	&\textbf{64.0\%}	&0.68	&\underline{0.089} \\
            \bottomrule[1pt]\midrule
        \end{tabular}%
    }
    \caption{Llama-3.2-3B-Instruct results across in-domain test set $D_{test}$, Agieval, Olympiad, and MMLU benchmarks. Among RL methods, the best values are highlighted in bold, and the suboptimal values are underlined. Our proposed C3RL method is shaded.}
    \label{tab:llama_calibration}
\end{table*}

\begin{figure}[h]
    \centering
    \includegraphics[clip, width=1\columnwidth]{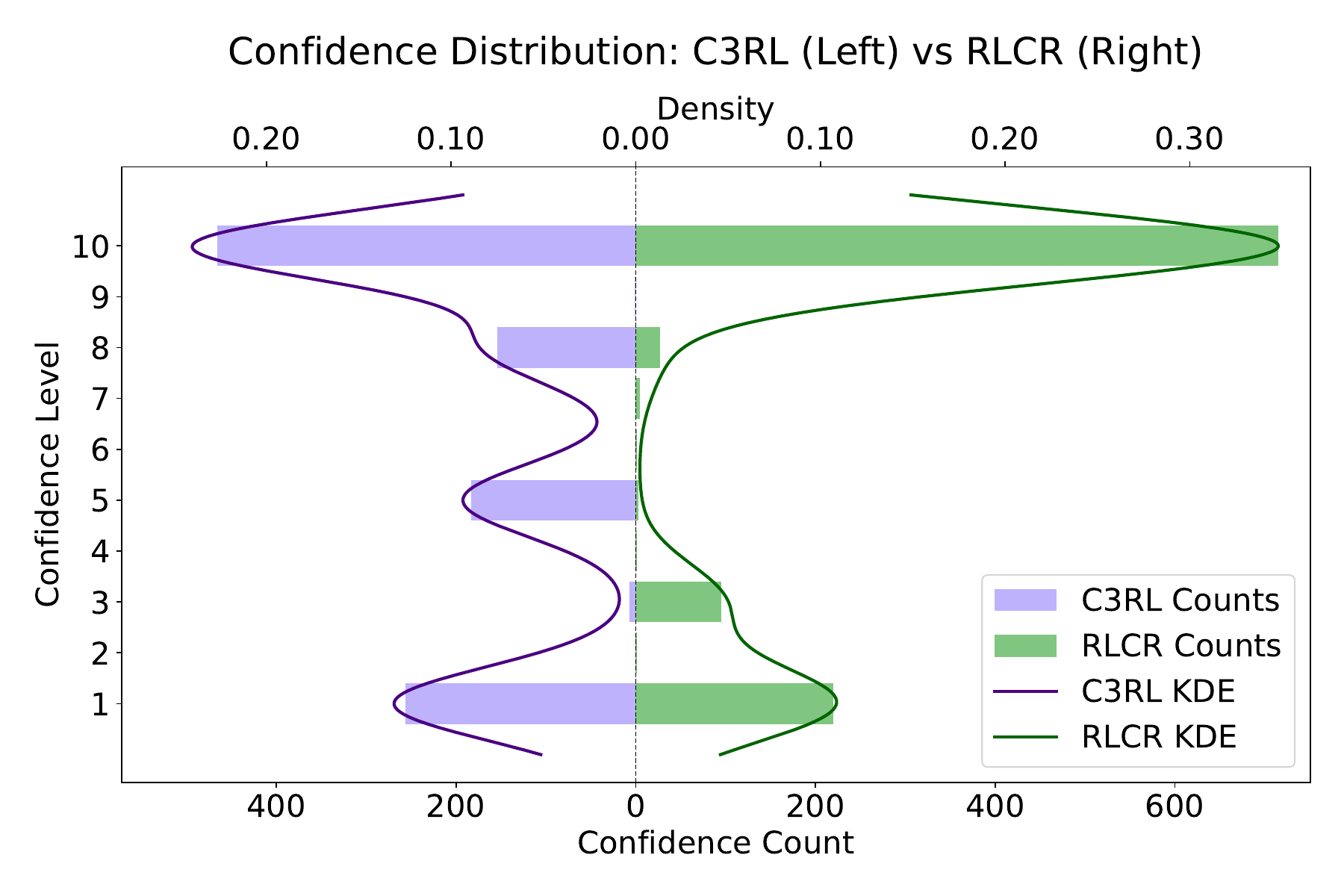}
    \caption{Confidence distribution (Count and Kernel Density Estimate) of C3RL and RLCR on in-domain test set $D_{test}$. RLCR confidence concentrates on $c=1$ and $c=10$, while C3RL confidence is more fine-grained to capture the nuanced uncertainty.}
    \label{fig:Conf distribution}
\end{figure}
\subsection{Confidence Distribution Comparison of C3RL and RLCR}
\label{appendix: confidence distribution}
As shown in Figure~\ref{fig:Conf distribution}, RLCR exhibits a highly polarized confidence distribution concentrated at the extremes ($c=1$ and $c=10$), effectively degenerating confidence estimation into a binary indicator. Although Brier-score-based training can align models on the semantic meaning of extreme confidence values (e.g., ``confidence=10'' denotes total certainty and ``confidence=1'' denotes total uncertainty), it does not provide a clear confidence boundary separating certain and uncertain cases. As a result, intermediate confidence levels are underutilized.

In contrast, C3RL explicitly introduces a confidence threshold ($t=5$) through threshold-guided calibration reward, encouraging the model to learn a clearer distinction between certainty and uncertainty while still preserving fine-grained confidence expression. Consequently, C3RL produces a substantially smoother confidence distribution with richer intermediate confidence values. This comparison highlights the importance of learning an explicit confidence threshold instead of relying solely on binary confidence extremes.

\subsection{Ablation Study to Determine $\alpha$ and $\beta$}
\label{appendix:ablation}
\begin{figure}[htbp]
    \centering
    \begin{subfigure}[b]{0.4\textwidth}
        \centering
        \includegraphics[width=\textwidth]{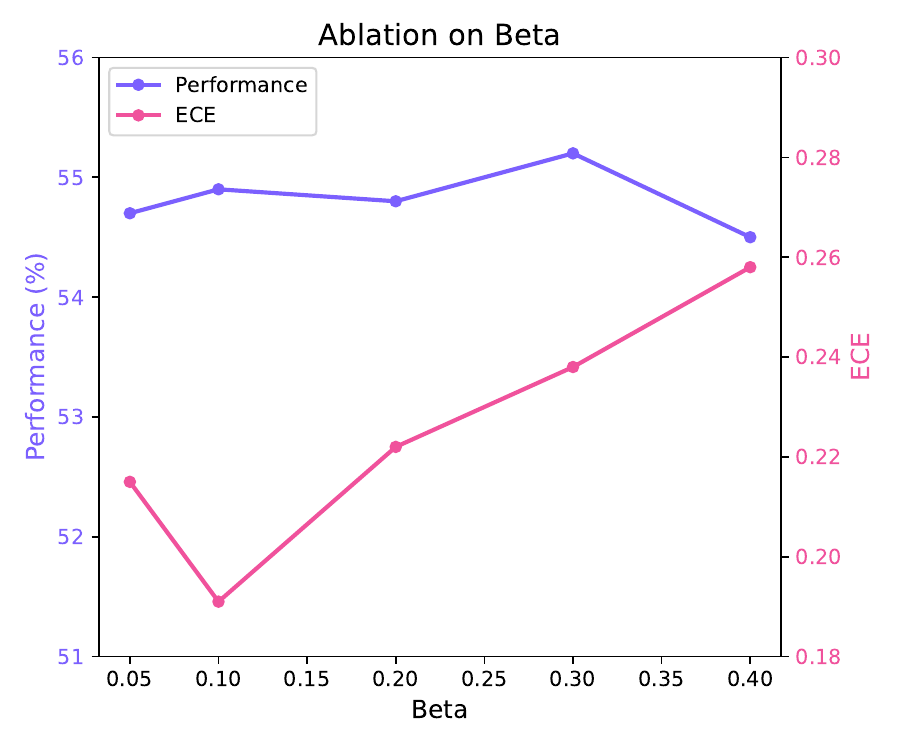} 
        \caption{Ablation of $\beta$} 
        \label{fig:beta}
    \end{subfigure}
    \\
    \begin{subfigure}[b]{0.4\textwidth}
        \centering
        \includegraphics[width=\textwidth]{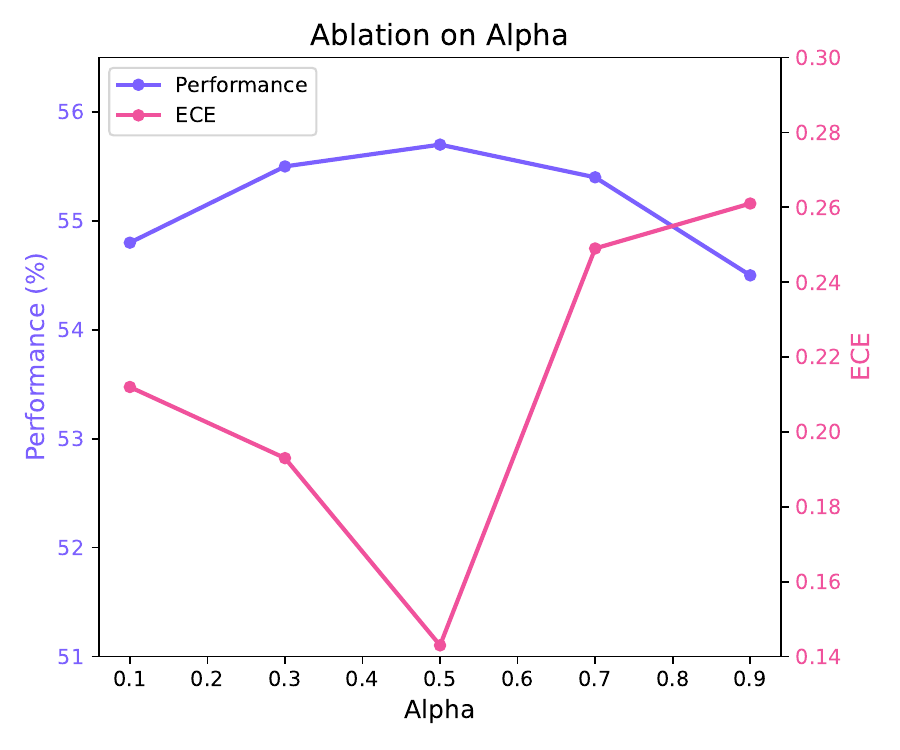}
        \caption{Ablation of $\alpha$ when $\beta=0.1$}
        \label{fig:alpha}
    \end{subfigure}
\end{figure}
We conduct ablation experiments to select optimal $\alpha$ and $\beta$. In these experiments, the correctness reward is always included, and the evaluation set is OOD (Text) specified in \ref{experiments: benchmarks_C3RL}. We first ablate $\beta$ by evaluating the models trained with different calibration rewards without reference accuracy reward. We try the $\beta$ values from [0.05, 0.1, 0.2, 0.3, 0.4] and the result is shown in Figure~\ref{fig:beta}. We find that different $\beta$ values share similar performance while $\beta=0.1$ gives notably drop in ECE, which stands for best calibration.

After fixing $\beta=0.1$, we then ablate the optimal $\alpha$ value from [0.1, 0.3, 0.5, 0.7, 0.9]. As shown in Figure~\ref{fig:alpha}, $\alpha=0.5$ reaches best performance and ECE at the same time, which displays its outstanding capability. So we choose $\beta=0.1$, $\alpha=0.5$ in our C3RL algorithm.

\subsection{Parallel-Inference Version of CAS}
\label{parallel CAS}
Assume there have been $n<N$ generated responses.
Based on the stopping mechanism, we can 
find the minimum additional inference time $k \in \{0, 1,...,N-n\}$ such that the probability will exceed the threshold:
\begin{equation}
\begin{split}
    \mathrm{Prob}(c_1 + k, c_2) &\geq P_{\mathrm{stop}} \\
    \mathrm{Prob}(c_1 + k - 1, c_2) &< P_{\mathrm{stop}} \notag
\end{split}
\end{equation}
Then $k$ additional responses are generated and $n$ is updated to $n+k$. When current stopping probability $\mathrm{Prob}(c_1,c_2)\geq P_{\mathrm{stop}}$ (i.e. $k=0$) or $n$ reaches the maximum inference time $N$, the inference ends and $a_{M}$ is returned as the final answer. The algorithm is described in Algorithm~\ref{alg:confidence_inference_parallel}.
\definecolor{codebg}{RGB}{255,255,255}
\begin{algorithm}
\caption{Confidence-based Adaptive Test Time Scaling (CAS) with Parallel Generation}
\label{alg:confidence_inference_parallel}
\begin{minipage}{\linewidth}
\colorbox{codebg}{%
\begin{minipage}{0.98\linewidth}
\begin{algorithmic}[1]
\Require Question $q$, max sampling time $N$, threshold $P_{\mathrm{stop}}$ 
\State Initialize $n \gets 0$, $k \gets 1$, $A_q \gets \emptyset$, $C_q \gets \emptyset$
\While{$n < N$ \textbf{and} stopping condition not met}
    \State \hangindent=2em \hangafter=1 Generate $k$ additional samples, $n \gets n+k$
    \State Update answer set $A_q$, confidence set $C_q$
    \State $c_1, c_2 \gets$ max and second max of $C_q$
    \State Compute $\mathrm{Prob} = \frac{1}{A} \int_0^{0.5} x^{c_2}(1-x)^{c_1} dx$
    \If{$\mathrm{Prob} \geq P_{\mathrm{stop}}$} \textbf{break} \EndIf
    \State \hangindent=8em \hangafter=0 {Find minimum $k$ for stopping criterion}
\EndWhile
\State \Return $a_M$ (answer with highest confidence sum $c_1$)
\end{algorithmic}
\end{minipage}%
}
\end{minipage}
\end{algorithm}

\subsection{Advantage of Verbalized Confidence}
We select verbalized confidence instead of internal logits or probablities to express uncertainty due to the following three reasons. First, verbalized confidence can express epistemic uncertainty of the responses, while the internal logits only represent uncertainty over tokens instead of the claim itself~\cite{lin2022teaching}. Second, verbalized confidence itself is typically better-calibrated than conditional probabilities~\cite{tian2023just}. Third, we want to explore a model-agnostic method, which should be applicable to all kinds of LLMs without relying on the internal state of LLMs~\cite{yang2024verbalized}. 

We test the efficiency of different uncertainty quantification methods in the task of mitigating CHOKE (Certain Hallucinations Overriding Known Evidence) examples ~\cite{trustmeimwrong}. As mentioned in~\cite{trustmeimwrong}, a CHOKE example needs to meet three requirements: (1) the LLM can originally answer the question correctly; (2) after modifying with the natural prompt, the LLM hallucinates; and (3) the LLM assigns a high certainty to the wrong answer. We conduct an evaluation experiment on the "all correct" + "partially correct" samples on the in-domain test set (808 among 1,073 samples). We use the model trained with C3RL to generate new responses following the identical settings in~\ref{appendix:evaluation settings}. The CHOKE examples are those with wrong answers after the prompt is modified (we follow the "prompt$\_$v7" modification in~\cite{trustmeimwrong} ) and assigned a high confidence. We use probability difference as a baseline. The detailed results are shown in Table~\ref{tab:choke}.
\begin{table}[htbp]
\centering
\renewcommand{\arraystretch}{0.6}
\begin{tabular}{p{2.4cm}p{1.8cm}p{1.8cm}}
\toprule
\textbf{Metric} & \textbf{Verb. Conf. + C3RL} & \textbf{Prob. Diff. + C3RL} \\
\midrule
\% of CHOKE & 6.1\% & 22.8\% \\
Avg. Conf. & 0.716 & 0.952 \\
\bottomrule
\end{tabular}
\caption{Performance comparison of verbalized confidence and probability difference on the task of reducing CHOKE examples.}
\label{tab:choke}
\end{table}

Table~\ref{tab:choke} shows the considerable advantage of verbalized confidence over probability difference method in decreasing CHOKE examples from 22.8\% to 6.1\%. Also, the extremely high average confidence of probability difference demonstrates it suffers from overconfidence, while verbalized confidence gives a relatively uniform confidence distribution.

\subsection{Selective-Answering Evaluation}

To demonstrate the effectiveness of confidence calibration in calibration-critical scenarios, we additionally evaluate our method on the selective-answering benchmark ALCUNA~\cite{alcuna}. ALCUNA contains both answerable and unanswerable questions. Specifically, the benchmark includes 88 answerable questions, where sufficient information exists to infer the correct answer, and 158 unanswerable questions, where entities are fabricated and no relevant knowledge exists.

For answerable questions, we report the same metrics as in the main experiments, including Accuracy, AUROC, and ECE. For unanswerable questions, we focus on the Abstain Rate, defined as:
\[
\text{Abstain Rate} = \frac{\# \text{abstentions}}{\# \text{samples}},
\]
where responses with confidence $\leq 5$ are regarded as abstentions.

Table~\ref{tab:alcuna_results} reports the evaluation results. As shown in the table, SaySelf achieves the highest abstention rate on unanswerable questions by aggressively suppressing confidence. However, this comes at the cost of severe degradation in both accuracy and calibration on answerable questions, indicating over-abstention even when answering is appropriate.

In contrast, C3RL achieves the best trade-off between accuracy and calibration on answerable samples while still maintaining a strong abstention capability on unanswerable samples. These results demonstrate that C3RL does not simply optimize abstention behavior, but instead learns a more reliable confidence boundary that distinguishes when to answer versus abstain.

\begin{table}[t]
\centering
\small
\begin{tabular}{lp{0.5cm}p{0.9cm}p{0.6cm}|p{1.8cm}}
\toprule
& \multicolumn{3}{c|}{Answerable} & Unanswerable \\
\cmidrule(r){2-4} \cmidrule(l){5-5}
Method & Acc$\uparrow$ & AUROC$\uparrow$ & ECE$\downarrow$ & Abstain Rate$\uparrow$ \\
\midrule
Base & 0.977 & 0.689 & 0.014 & 0.247 \\
RLVR & 0.966 & 0.500 & 0.034 & 0.234 \\
SaySelf & 0.966 & \textbf{0.859} & 0.242 & \textbf{0.962} \\
RLCR & \textbf{0.987} & 0.483 & 0.015 & 0.665 \\
C3RL w/o Ref & \textbf{0.989} & 0.454 & 0.024 & 0.595 \\
C3RL (Ours) & 0.977 & 0.727 & \textbf{0.002} & 0.734 \\
\bottomrule
\end{tabular}
\caption{Evaluation results on the ALCUNA selective-answering benchmark. For answerable questions, we report Accuracy, AUROC, and ECE. For unanswerable questions, we report Abstain Rate, where responses with confidence $\leq 5$ are treated as abstentions.}
\label{tab:alcuna_results}
\end{table}

\subsection{Evaluation Results on Canonical QA Benchmarks}

To evaluate the generalization capability of C3RL beyond reasoning-intensive tasks, we additionally conduct experiments on two canonical question-answering benchmarks: CommonSenseQA~\cite{commonsenseqa} and StrategyQA~\cite{strategyqa}. These datasets mainly focus on commonsense reasoning and implicit world knowledge, which are substantially different from our training corpus constructed from mathematics, science, and logical reasoning datasets.

As shown in Table~\ref{tab:canonical_qa_results}, C3RL consistently achieves the best accuracy-calibration trade-off across both benchmarks. On CommonSenseQA, C3RL achieves comparable accuracy to RLCR while significantly reducing calibration error. On StrategyQA, C3RL surpasses all baselines in both accuracy and AUROC. These results demonstrate that the effectiveness of threshold-guided calibration generalizes beyond the reasoning-intensive training distribution and remains beneficial on canonical QA tasks.

\begin{table}[t]
\centering
\small
\renewcommand{\arraystretch}{0.6}
\begin{tabular}{
p{1cm}
>{\centering\arraybackslash}p{0.4cm}
>{\centering\arraybackslash}p{0.9cm}
>{\centering\arraybackslash}p{0.6cm}|
>{\centering\arraybackslash}p{0.4cm}
>{\centering\arraybackslash}p{0.9cm}
>{\centering\arraybackslash}p{0.6cm}
}
\toprule
& \multicolumn{3}{c|}{CommonSenseQA} & \multicolumn{3}{c}{StrategyQA} \\
\cmidrule(r){2-4} \cmidrule(l){5-7}
Method & Acc$\uparrow$ & AUROC$\uparrow$ & ECE$\downarrow$
& Acc$\uparrow$ & AUROC$\uparrow$ & ECE$\downarrow$ \\
\midrule
SaySelf & 0.725 & \textbf{0.721} & 0.183
& 0.496 & 0.589 & \textbf{0.331} \\
RLCR & \textbf{0.786} & 0.604 & 0.197
& 0.603 & 0.588 & 0.359 \\
C3RL & 0.785 & 0.608 & \textbf{0.093}
& \textbf{0.643} & \textbf{0.594} & 0.356 \\
\bottomrule
\end{tabular}
\caption{Evaluation results on canonical QA benchmarks CommonSenseQA and StrategyQA.}
\label{tab:canonical_qa_results}
\end{table}

\subsection{Ablation Experiment Evaluating RLCR + Reference Accuracy Reward}
\label{appendix:rlcr_ref_ablation}

We additionally conduct an ablation experiment by incorporating the reference accuracy reward into RLCR, denoted as \textbf{RLCR w.t. Ref}. This experiment aims to verify whether the improvement of C3RL mainly comes from introducing additional reward terms or from the threshold-guided calibration formulation itself.

Table~\ref{tab:rlcr_ref_ablation} reports the evaluation results on AGIEval and MMLU. Compared with RLCR, directly adding the reference accuracy reward to RLCR does not improve calibration performance and even slightly degrades AUROC and ECE on both benchmarks. In contrast, C3RL w/o Ref consistently outperforms RLCR and RLCR w.t. ref in calibration metrics, while the full C3RL achieves the best overall accuracy-calibration trade-off.

These results demonstrate that the improvement of C3RL does not simply arise from adding extra reward signals. Instead, the threshold-guided calibration objective plays the dominant role in preventing confidence collapse and enabling more fine-grained uncertainty estimation.

\begin{table}[t]
\centering
\small
\begin{tabular}{
p{1.77cm}
>{\centering\arraybackslash}m{0.35cm}
>{\centering\arraybackslash}m{0.75cm}
>{\centering\arraybackslash}m{0.55cm}|
>{\centering\arraybackslash}m{0.35cm}
>{\centering\arraybackslash}m{0.75cm}
>{\centering\arraybackslash}m{0.5cm}
}
\toprule
& \multicolumn{3}{c|}{AGIEval} & \multicolumn{3}{c}{MMLU} \\
\cmidrule(r){2-4} \cmidrule(l){5-7}
Method & Acc$\uparrow$ & AUROC$\uparrow$ & ECE$\downarrow$
& Acc$\uparrow$ & AUROC$\uparrow$ & ECE$\downarrow$ \\
\midrule
SaySelf & 0.510 & \textbf{0.746} & \textbf{0.148}
& 0.642 & \textbf{0.732} & 0.128 \\
RLCR & 0.577 & 0.663 & 0.203
& 0.709 & 0.607 & 0.174 \\
RLCR w.t.Ref & 0.570 & 0.647 & 0.214
& 0.710 & 0.584 & 0.190 \\
C3RL w/oRef & \textbf{0.580} & 0.684 & 0.198
& 0.712 & 0.653 & 0.111 \\
C3RL (Ours) & 0.577 & 0.694 & 0.154
& \textbf{0.716} & 0.652 & \textbf{0.089} \\
\bottomrule
\end{tabular}
\caption{Evaluation results on AGIEval and MMLU for RLCR with additional reference accuracy reward (RLCR w.t. Ref). The results show that simply adding the reference reward to RLCR does not recover the advantage of C3RL, demonstrating the effectiveness of threshold-guided calibration.}
\label{tab:rlcr_ref_ablation}
\end{table}

\subsection{AI Using Statement}
We employ copilot in writing code for experiments and drawing graphs. We guarantee that all the materials have been checked carefully manually.

\subsection{Case Presentation}
\begin{table*}[t]
\small
\centering
\scalebox{1}{
\begin{tcolorbox}[colback=gray!00,
                  colframe=black,
                  width=16.5cm,
                  arc=1.5mm, auto outer arc,
                  breakable,
                  left=0.9mm, right=0.9mm,
                  boxrule=0.9pt, colbacktitle = black!65!black,
                  title = {Cases of different RL methods in calibration}
                 ]
{\textbf{Query:} The muscles of the soft palate are innervated by branches of the A) trigeminal and glossopharyngeal nerves. B) facial and glossopharyngeal nerves. C) trigeminal and vagus nerves. D) facial and vagus nerves.
} \\
{\textbf{Ground Truth:} C.
}
\vspace{1pt}
\textbf{C3RL Model response:} 
                    To determine the correct answer, let's analyze the function and innervation of the muscles of the soft palate.\textbackslash n\textbackslash n1. The soft palate is a part of the oral cavity that helps in various functions such as swallowing, speech, and maintaining airway closure during breathing.\textbackslash n2. The muscles of the soft palate are innervated by branches of specific cranial nerves.\textbackslash n\textbackslash nNow, let's consider each option:\textbackslash n- **A) trigeminal and glossopharyngeal nerves**: The trigeminal nerve (CN V) innervates the sensory receptors in the mouth and throat, but it does not innervate the muscles of the soft palate. The glossopharyngeal nerve (CN IX) does innervate the muscles of the soft palate.\textbackslash n- **B) facial and glossopharyngeal nerves**: The facial nerve (CN VII) innervates the muscles of the face but not the soft palate. The glossopharyngeal nerve (CN IX) does innervate the muscles of the soft palate.\textbackslash n- **C) trigeminal and vagus nerves**: The trigeminal nerve (CN V) does not innervate the muscles of the soft palate. The vagus nerve (CN X) does innervate the muscles of the soft palate.\textbackslash n- **D) facial and vagus nerves**: The facial nerve (CN VII) does not innervate the muscles of the soft palate. The vagus nerve (CN X) does innervate the muscles of the soft palate.\textbackslash n\textbackslash nFrom the above analysis, the correct innervation for the muscles of the soft palate is by the glossopharyngeal nerve (CN IX). \colorbox{green!30}{<ANSWER is: $\backslash$boxed\{A\}>|<CONFIDENCE:1>}.

\textbf{Base model response:} 
                    To determine which nerves innervate the muscles of the soft palate, we need to consider the anatomical innervation of this area.\textbackslash n\textbackslash n1. **Trigeminal Nerve (CN V)**: This nerve provides sensory function to the face and oral cavity. The branch involved in the soft palate is the greater petrosal nerve, which connects to the facial nerve (CN VII).\textbackslash n\textbackslash n2. **Glossopharyngeal Nerve (CN IX)**: This nerve provides both sensory and motor functions. For the soft palate, it innervates the tensor veli palatini muscle, which is essential for palate movement during swallowing.\textbackslash n\textbackslash n3. **Facial Nerve (CN VII)**: This nerve provides motor function to the muscles of facial expression and also innervates the tensor veli palatini muscle via the greater petrosal nerve.\textbackslash n\textbackslash n4. **Vagus Nerve (CN X)**: This nerve has extensive distribution throughout the body and does not specifically innervate the soft palate muscles.\textbackslash n\textbackslash nGiven the above points, the muscles of the soft palate are primarily innervated by the glossopharyngeal nerve and the facial nerve, through their connection with the greater petrosal nerve.\textbackslash n\textbackslash n\colorbox{red!30}{<ANSWER is: $\backslash$boxed\{B\}|<CONFIDENCE:8>}.

\textbf{RLVR Model response:}
                    To determine the correct answer, we need to identify which nerves innervate the muscles of the soft palate. The soft palate is part of the pharynx and is involved in various functions such as swallowing and speech.\textbackslash n\textbackslash n1. **Trigeminal Nerve (CN V)**: This nerve innervates the sensation of the skin on the face and the muscles of mastication. It does not innervate the muscles of the soft palate.\textbackslash n\textbackslash n2. **Glossopharyngeal Nerve (CN IX)**: This nerve innervates the muscles of the soft palate, the pharynx, and the tonsils. It also carries taste fibers from the posterior third of the tongue.\textbackslash n\textbackslash n3. **Facial Nerve (CN VII)**: This nerve innervates the muscles of facial expression but does not innervate the muscles of the soft palate.\textbackslash n\textbackslash n4. **Vagus Nerve (CN X)**: This nerve innervates the larynx, the heart, and the muscles of the digestive tract. It does not innervate the muscles of the soft palate.\textbackslash n\textbackslash nGiven this information, the muscles of the soft palate are innervated by branches of the glossopharyngeal nerve (CN IX) and the vagus nerve (CN X).\textbackslash n\textbackslash nHowever, the correct choice that matches the options provided is:\textbackslash n\textbackslash nB) facial and glossopharyngeal nerves.\textbackslash n\textbackslash nBut since the correct choice should be B, the answer is: \colorbox{red!30}{<ANSWER is: $\backslash$boxed\{B\}>|<CONFIDENCE:8>}.

\textbf{SaySelf Model response:}
                    <think> The muscles of the soft palate are innervated by branches of the trigeminal nerve (CN V) and the glossopharyngeal nerve (CN IX). The trigeminal nerve innervates the anterior two-thirds of the tongue and the soft palate, while the glossopharyngeal nerve innervates the posterior one-third of the tongue and the soft palate. Therefore, the correct answer is A.\textbackslash n\textbackslash n</think>\textbackslash n\colorbox{red!30}{<ANSWER is: $\backslash$boxed\{A\}>|<CONFIDENCE:10>}.

\textbf{RLCR Model response:} 
                    <think> The muscles of the soft palate are innervated by branches of the trigeminal and glossopharyngeal nerves. The trigeminal nerve (CN V) innervates the musculature of the soft palate through the greater palatine nerve, and the glossopharyngeal nerve (CN IX) innervates it through the lesser palatine nerves. Therefore, the correct answer is A.</think>\textbackslash n \colorbox{red!30}{<ANSWER is: $\backslash$boxed\{A\}>|<CONFIDENCE:10>}.

\end{tcolorbox}
}

\caption{Responses of C3RL model, Base model, RLVR model, SaySelf model and RLCR model on a MMLU question}
\label{tab:C3RL_case}
\end{table*}

We present a case from MMLU anatomy questions in Table~\ref{tab:C3RL_case}. All the responses are incorrect, while only the model trained by C3RL signals a low confidence, showing the effectiveness of C3RL in calibration.

\end{document}